\documentclass{article}

\usepackage{microtype}
\usepackage{graphicx}
\usepackage{booktabs} 

\usepackage[utf8]{inputenc}
\usepackage[cmex10]{amsmath}
\usepackage{mathtools}
\usepackage{amsfonts}
\usepackage{xspace}
\usepackage{tikz}
\usepackage{comment}
\usepackage{subcaption}

\newcommand{\asyncfl}{{AsyncFL}\xspace}
\newcommand{\syncfl}{{SyncFL}\xspace}
\newcommand{\syncflos}{{SyncFL with over-selection}\xspace}
\newcommand{\syncflnoos}{{SyncFL without over-selection}\xspace}

\newcommand{\papaya}{{\sc Papaya}\xspace}

\usepackage{hyperref}

\usepackage[accepted]{mlsys2022}

\mlsystitlerunning{\papaya: Practical, Private, and Scalable Federated Learning}

\usepackage[capitalize]{cleveref}
\crefname{step}{step}{steps}
\crefname{figure}{Figure}{Figures}

\begin{document}

\twocolumn[
\mlsystitle{\papaya: Practical, Private, and Scalable Federated Learning}



\mlsyssetsymbol{equal}{*}

\begin{mlsysauthorlist}
\mlsysauthor{Dzmitry Huba}{fb}
\mlsysauthor{John Nguyen}{fb}
\mlsysauthor{Kshitiz Malik}{fb}
\mlsysauthor{Ruiyu Zhu}{fb}
\mlsysauthor{Mike Rabbat}{fb}
\mlsysauthor{Ashkan Yousefpour}{fb}
\mlsysauthor{Carole-Jean Wu}{fb}
\mlsysauthor{Hongyuan Zhan}{fb}
\mlsysauthor{Pavel Ustinov}{fb}
\mlsysauthor{Harish Srinivas}{fb}
\mlsysauthor{Kaikai Wang}{fb}
\mlsysauthor{Anthony Shoumikhin}{fb}
\mlsysauthor{Jesik Min}{fb}
\mlsysauthor{Mani Malek}{fb}
\end{mlsysauthorlist}

\mlsysaffiliation{fb}{Meta AI, USA}

\mlsyscorrespondingauthor{Dzmitry Huba}{huba@fb.com}
\mlsyscorrespondingauthor{John Nguyen}{ngjhn@fb.com}
\mlsyscorrespondingauthor{Kshitiz Malik}{kmalik2@fb.com}

\mlsyskeywords{Machine Learning, MLSys}

\vskip 0.3in

\begin{abstract}
Cross-device Federated Learning (FL) is a distributed learning paradigm with several challenges that differentiate it from traditional distributed learning, variability in the system characteristics on each device, and millions of clients coordinating with a central server being primary ones. Most FL systems described in the literature are synchronous -- they perform a synchronized aggregation of model updates from individual clients. Scaling synchronous FL is challenging since increasing the number of clients training in parallel leads to diminishing returns in training speed, analogous to large-batch training. Moreover, stragglers hinder synchronous FL training. In this work, we outline a production asynchronous FL system design. Our work tackles the aforementioned issues, sketches of some of the system design challenges and their solutions, and touches upon principles that emerged from building a production FL system for millions of clients. Empirically, we demonstrate that asynchronous FL converges faster than synchronous FL when training across nearly one hundred million devices. In particular, in high concurrency settings, asynchronous FL is 5$\times$ faster and has nearly 8$\times$ less communication overhead than synchronous FL. 

\end{abstract}
]



\printAffiliationsAndNotice{} 

\section{Introduction}
\label{sec:intro}
Cross-device \emph{federated learning} (FL) is a distributed learning paradigm where a large collection of clients collaborate to train a machine learning model while the raw training data stays on client devices. FL promises to train high-quality models by leveraging data from massive client populations, while ensuring security and privacy of client data.

In traditional parallel systems, \emph{concurrency} refers to the number of processors running a parallel application, and \emph{utilization} refers to the fraction of processors actively computing at any time. In this paper we focus on the scalability of FL systems: \textit{``a measure of [their] capacity to effectively utilize an increasing number of processors''} \cite{scalability}. 
In the context of FL, concurrency refers to the number of clients training simultaneously, and our aim is to develop systems that can accelerate training by training concurrently on more clients. Companies like Apple, Meta, Google, and others have the potential to scale FL training to \textit{hundreds of millions or billions of clients}. 

\begin{figure}[t]
     \centering
     \includegraphics[width=0.9\columnwidth]{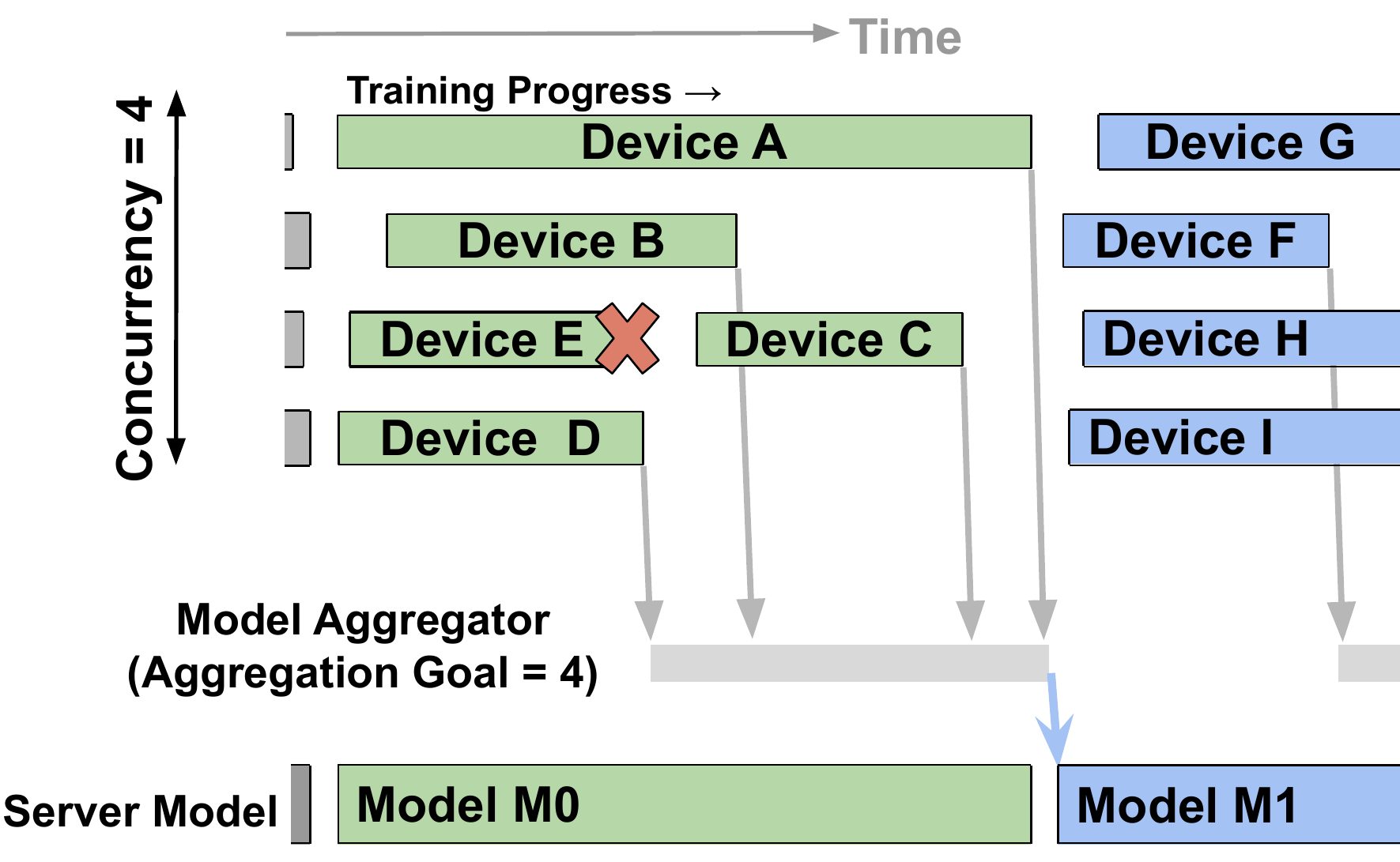}
     \caption{Example of SyncFL with a concurrency of 4; i.e., up to 4 devices can train in parallel. The server model is updated once all clients are ready (once the \emph{aggregation goal} is achieved), so concurrency determines how frequently a new server model is produced. If one client drops out mid-round (e.g., Device E), a new client, (e.g., Device C) is selected to take its place. Utilization decreases once some devices have returned updates, and the round completion time depends on the slowest-returning device --- the straggler. To overcome stragglers, \emph{over-selection} is often used, in which case the concurrency may be higher than the aggregation goal, and once the aggregation goal is achieved, updates from other devices still processing are discarded. Note that mid-round client replacement, like Device C replacing Device E, is not possible in some other SyncFL Systems~\cite{google-fl}, but is possible in \papaya's implementation of SyncFL. We see up to 10\% of clients drop.}
     \label{fig:sync_timeline}
\end{figure}

Prior work describing FL systems has focused on synchronous training~\cite{google-fl,apple-fl,ibm-fl-full,clara,fate}. Synchronous FL (\syncfl) training proceeds in rounds, as illustrated in Figure~\ref{fig:sync_timeline}. The number of clients participating in each round corresponds to the concurrency.\footnote{In the SyncFL literature, concurrency is also referred to as \textit{clients per round}~\cite{google-fedavg}.} In each round, clients download the current server model, train this model locally on their respective data, and report a model update back to the server. Once all client updates are ready, they are aggregated, and then the server computes the new model using the aggregated updates.


SyncFL faces two main challenges when scaling. First, there are many sources of heterogeneity in cross-device FL~\cite{fl-survey}: clients have different hardware capabilities (processor speeds, memory sizes), and data can be highly imbalanced across clients, with some clients having multiple orders of magnitude more data than others. In synchronous systems, heterogeneity results in \emph{stragglers} --- clients in the tail take much longer to complete local training and prolong the time to complete each round of training, hampering utilization. Over-selection is commonly used to reduce the impact of stragglers on the runtime of SyncFL methods~\cite{google-fl}. Over-selection results in discarding updates from the slowest-responding clients selected in each round, and it has been noted that this may bias the trained model against slow-responding clients.

The second challenge SyncFL faces is that increasing concurrency in synchronous training corresponds to using larger cohorts (group of clients participating in a round); i.e., more user updates averaged before performing a server update. This leads to similar effects as using large batches in traditional data-parallel training~\cite{keskar2017large}. Large-cohort training has been found to make inefficient use of client updates~\cite{google-fl,charles2021large}. Consequently, increasing cohort size does not reduce wall-clock training time proportionally.


Asynchronous FL (\asyncfl, see Section~\ref{sec:proposed_design}) 
can potentially alleviate these challenges. In \asyncfl, clients return updates to be aggregated as soon as the updates are ready, and a new client may then begin computing updates immediately. Client training is decoupled from server model updates. Consequently, \asyncfl is not impacted by stragglers and utilization can be kept high (essentially at 100\%) throughout training. However, as with all asynchronous systems, \asyncfl must handle \emph{staleness} --- updates from clients, especially slow-responding clients, based on a server model that has been updated many times in the interim, and hence may not provide useful information for training~\cite{bertsekasTsitsiklis}. AsyncFL methods have been previously explored~\cite{fedasync,fedbuff,xu2021asynchronous}, but none has yet been demonstrated and evaluated at scale.


\textbf{Contributions.} This paper presents \papaya,\footnote{Why \papaya? Say ``privacy-preserving AI'' five times in a row, fast.} \textbf{\textit{the first production FL system to support asynchronous and synchronous training at scale.}} We introduce a novel asynchronous secure aggregation protocol, allowing clients to communicate updates to the server in a cryptographically secure manner without needing to wait until other clients are ready to perform secure aggregation. This enables the implementation of FL with buffered asynchronous aggregation that has been recently introduced in~\cite{fedbuff}. 

We evaluate \papaya in Section~\ref{sec:evaluation} by training a language model for next-word prediction on a population of \textit{millions} of devices in the field. We demonstrate that AsyncFL is substantially more scalable than SyncFL. 
Although asynchronous execution results in some stale client responses, staleness in AsyncFL can be controlled by choosing an appropriate \emph{aggregation goal} in buffered asynchronous aggregation~\cite{fedbuff}. The aggregation goal is the number of client updates that need to be received before the server performs a model update. 
Consequently, AsyncFL can compute many more server updates than SyncFL in a fixed amount of time, leading to much better scaling than SyncFL. Moreover, with AsyncFL, the number of server updates per unit time increases nearly linearly with concurrency. When comparing both approaches in terms of wall-clock time to reach a target test loss, \textbf{\textit{we show that AsyncFL is 
almost 5$\times$ faster  
and 8$\times$ more communication-efficient than SyncFL.}}

Finally, \textbf{\textit{we demonstrate that AsyncFL achieves more fair models than SyncFL with over-selection}}. We observe very high correlation between slow devices and devices with many training samples. Discarding the updates from slow devices results in biasing the model trained using SyncFL with over-selection: the test perplexity for clients in the 99th percentile increases by 53\% when enabling over-selection. This bias is not introduced when training with AsyncFL.

\section{Understanding the landscape of federated learning at-scale}
\label{sec:landscape}
\begin{figure}[t]
     \centering
     \begin{minipage}[t]{\linewidth}
         \centering
         \includegraphics[width=\linewidth]{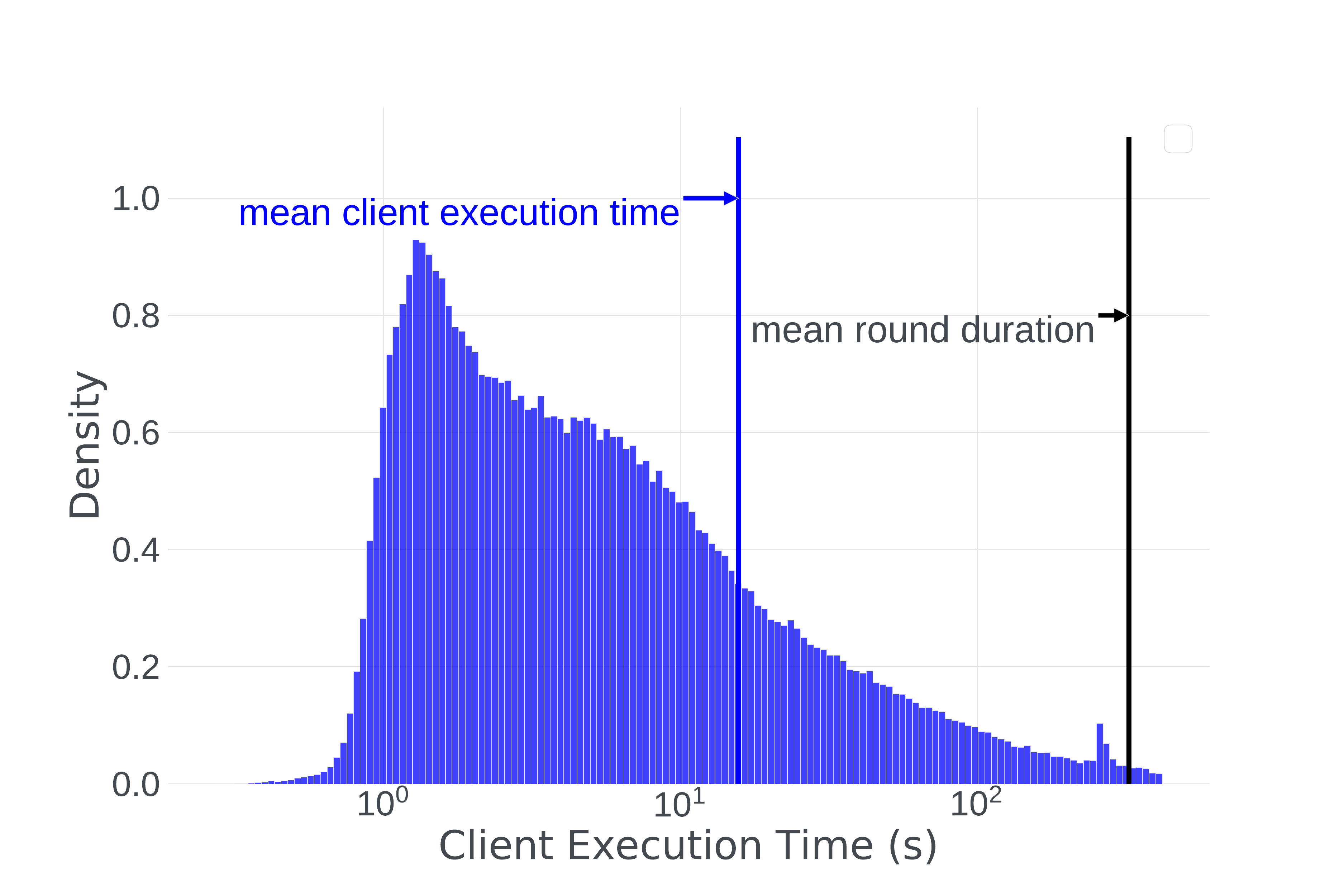}
     \end{minipage}
     
     \caption{Histogram of client execution times (note: x-axis is on a logarithmic scale). Because of stragglers, the mean round duration of SyncFL with concurrency set to 1000 is much larger than the mean client execution time.} 
     \label{fig:sync_client_execution}

\end{figure}

Building a robust federated learning system faces key design challenges:
\begin {itemize}

\item \textit{System and data heterogeneity}, where client devices participating in FL exhibit different system characteristics and possess different amounts of training data, leading to large differences in training time, and
\item \textit{Scalability}, where the training time speedup with higher degree of concurrency experiences diminishing return and plateaus quickly. 
\end{itemize}

To demonstrate the impact of the aforementioned challenges faced by FL, we begin by examining the degree of data and system heterogeneity observed in production when \textit{hundreds of millions} of client devices jointly train a global model. To understand the limit of SyncFL approaches, we take a data-driven approach to demonstrate the impact of \textit{scale} on the state-of-the-art synchronous model aggregation protocol.

\textbf{System and Data Heterogeneity.} 
Compute capabilities of mobile devices in the field differ by an order of magnitude ~\cite{inference-at-the-edge}. Moreover, the number of training examples also varies widely across users~\cite{leaf}. In combination, system and data heterogeneity can result in large differences in training time. Variance in training time results in \textit{stragglers} that slow down the overall training time in \syncfl.

Figure~\ref{fig:sync_client_execution} shows the distribution of training times across millions of clients for a common FL application (language model training, Section~\ref{sec:evaluation}). The per-client training time distribution spans more than two orders of magnitude. When running \syncfl with concurrency and aggregation goal set to 1000, the average round completion time is $21\times$ larger than the mean client training time. 

To mitigate the impact of stragglers in \syncfl, some systems use over-selection~\cite{google-fl}. In Section~\ref{sec:sampling_bias}, we show that over-selection causes sampling bias, thus producing models that are unfair to stragglers.




\begin{figure}[t]
     \centering
     \begin{minipage}[t]{\linewidth}
         \centering
         \includegraphics[width=\columnwidth]{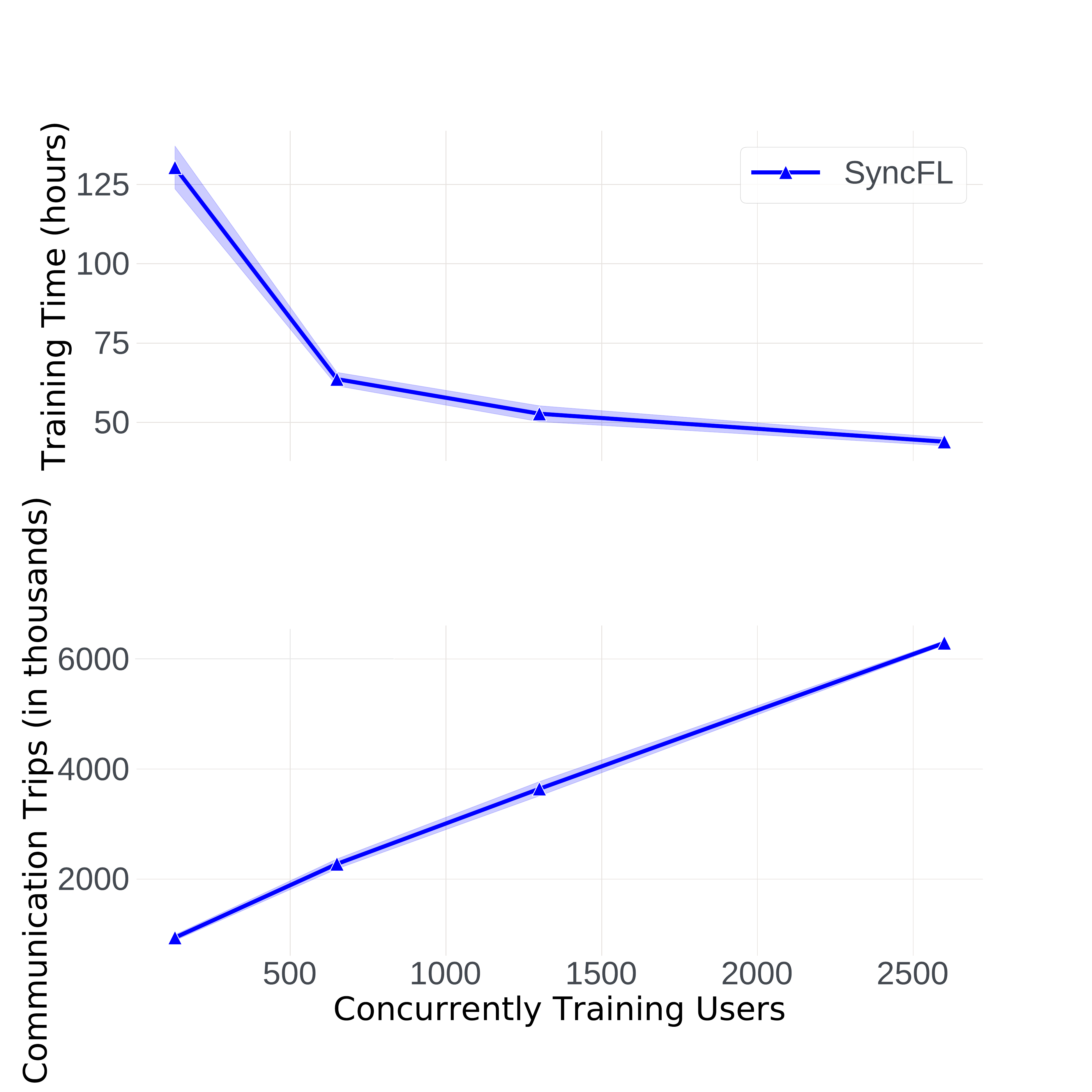}
     \end{minipage}
     \caption{
     We use \syncfl to train a language model until it reaches a target accuracy, while varying the concurrency from 130 to 2600 in Section~\ref{sec:evaluation}. The client population is around 100 million
     (Top) As concurrency increases, training time decreases rapidly at first, but then plateaus. (Bottom) As concurrency increases, \syncfl becomes communication inefficient. Communication trips refers to the number of client updates received at the server.}
     \label{fig:sync_scalability}
\end{figure}

\textbf{Scalability.} To further minimize the training time to convergence, a straightforward approach is to scale up the overall training throughput of the FL system by increasing the degree of training concurrency. 
Figure~\ref{fig:sync_scalability} illustrates the training time to convergence and the communication overhead for the \syncfl method FedAdam~\cite{adaptive-fl-optimization} as the number of concurrently training users increases from 130 to 2600. 
As concurrency increases, training time decreases slowly, while communication resource consumption increases much faster. For example, doubling the concurrency from 1300 to 2600 decreases the overall training time by only 17\% while increasing communication costs by 73\%.

We need resilient solutions that handle \textit{data} and \textit{system heterogeneity} at scale. At the same time, as shown in Figure~\ref{fig:sync_scalability}, we are at the scaling limit of synchronous model aggregation. To build FL suitable for billions of clients, we need a fundamentally different model aggregation protocol that is resilient to heterogeneity (client independence), scalable to large cohort sizes (beyond the order of hundreds), and secure (asynchronous secure aggregation).  
Next we describe the proposed design of \papaya and demonstrate how \asyncfl can improve large-scale FL by improving scalability and straggler resilience.



\section{Proposed Design}
\label{sec:proposed_design}
In this section, we first describe the \asyncfl algorithm \papaya uses. Next, we discuss the challenges in implementing \asyncfl in a large-scale production system.

\subsection {AsyncFL Algorithm}

\begin{figure}[t]
     \centering
     \includegraphics[width=0.9\columnwidth]{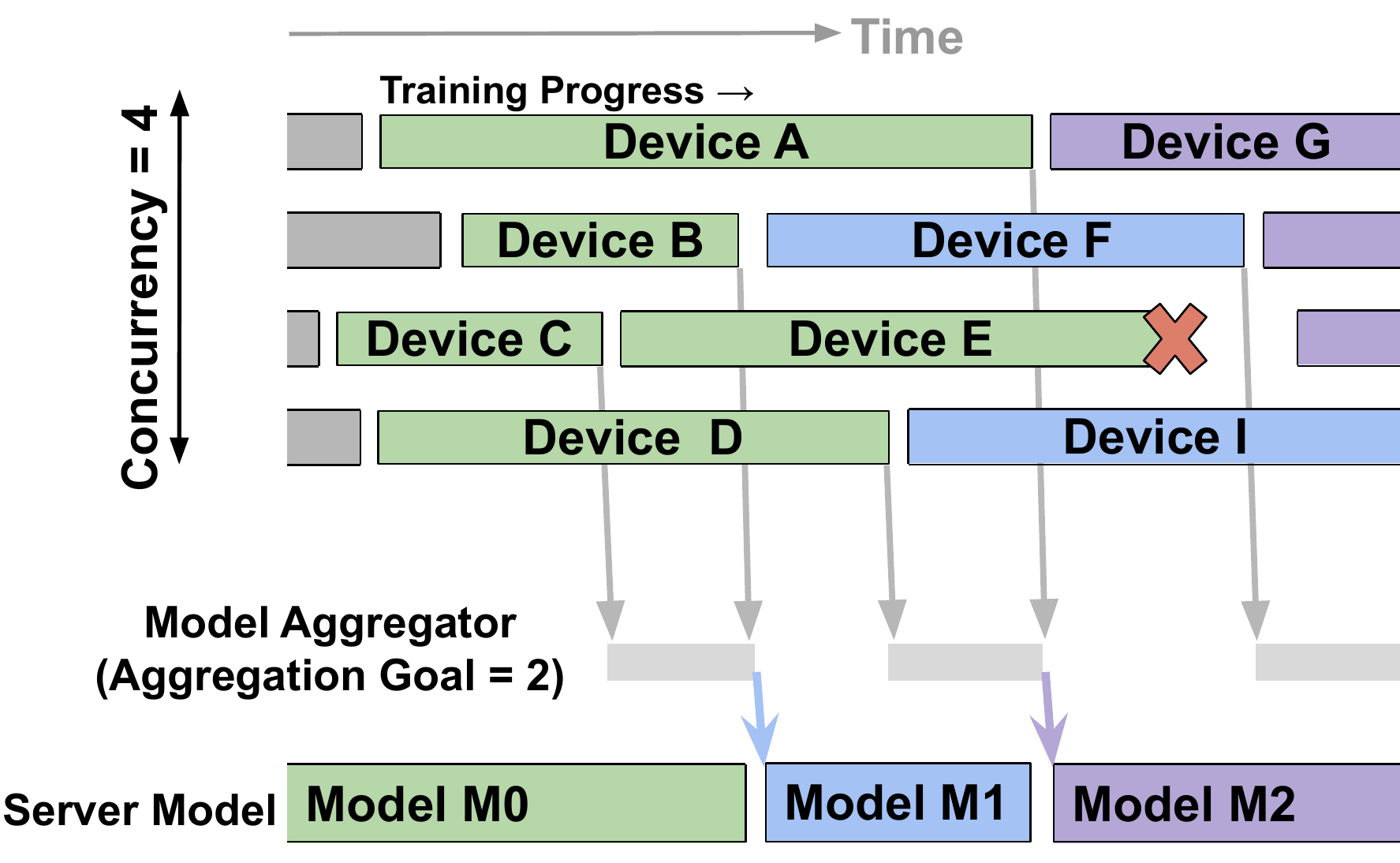}
     \caption{Example of AsyncFL with concurrency of 4, and where the aggregation goal is 2. In SyncFL with over-selection, the aggregation goal is less than the concurrency to reduce the straggler effect, but this results in wasted client effort and can lead to model bias. In contrast, running AsyncFL with aggregation goal less than concurrency does not result in wasted client effort, but rather some updates may be \emph{stale}. However, staleness can be controlled by increasing the aggregation goal, and utilization remains high throughout training.}
      \label{fig:async_timeline}
\end{figure}

\papaya implements a recently proposed AsyncFL algorithm, FedBuff~\cite{fedbuff}. In FedBuff, there is no notion of rounds: clients download, train, and upload updates asynchronously (Figure~\ref{fig:async_timeline}). After a client finishes local training, it uploads the model update (difference between the trained local model and initial model it received from the server before training).  
The aggregator tracks progress towards an \emph{aggregation goal}, the number of client updates that need to be received before the server performs a model update.
As soon as the aggregation goal has been achieved, the aggregated update is released and the server model update is performed.
Each client update is weighted by the number of examples the client trained on and a factor depending on the staleness of the update. \emph{Staleness} is defined as the difference between the model version that a client uses to start local training and the server model version at the time when a client uploads its model update. For example, Figure~\ref{fig:async_timeline} shows FedBuff with 4 concurrent users and an aggregation goal of 2. Device A's update has a staleness of 1 since the server model was updated once while Device~A was training.
In the rest of the paper, \asyncfl refers to our implementation of the FedBuff algorithm in \papaya.

We show in Section~\ref{sec:evaluation} that in a large-scale production setting with system and data heterogeneity, \asyncfl is faster and more resource efficient than \syncfl. 
However, \asyncfl brings a unique set of challenges that require careful system design.

\subsection {System Design challenges in AsyncFL}
Existing large-scale FL systems are designed to run \syncfl~\cite{google-fl, apple-fl}. Hence, their architectures are not compatible with asynchronous training. There are four main reasons for this incompatibility, which we discuss next.

\textbf{Client Selection. } Client selection in \syncfl is based on forming synchronous cohorts. For example, in~\citet{google-fl} a client cannot begin training until the entire cohort of clients has been selected. To support \asyncfl, we design a client selection mechanism that avoids any inter-client dependencies (Section~\ref{sec:client_independence}).

\textbf{Secure Aggregation. } Secure Aggregation (SecAgg) improves the privacy of FL algorithms by hiding individual client model updates ensuring that the server can only view the final aggregation of all model updates. Most FL systems implement SecAgg based on secure multi-party computation (SMPC)~\cite{secagg, so2021turbo}. SMPC-based SecAgg requires clients participating in a round to form a cohort and run a multi-leg protocol through the duration of the round. These requirements are not compatible with asynchronous training. 

Motivated by these challenges, we propose a novel incremental Asynchronous Secure Aggregation algorithm that uses a Trusted Execution Environment ~\cite{secagg-sgx} in Section~\ref{sec:async_secagg}. 

\textbf{Client Replacement for High Utilization. } Cohort-based \syncfl systems do not replace clients in the middle of a round~\citep{google-fl}. However, \asyncfl requires continuous replacement of clients that have finished training or have failed. We describe a fast client replacement mechanism that enables our \asyncfl implementation to achieve close to 100\% client utilization, significantly higher than \syncfl(Section~\ref{sec:sustained_high_concurrency}). 

\textbf{Support for Fast Model Aggregation. } \asyncfl generates up to 30$\times$ more server model updates per unit time than \syncfl, as shown below in Figure~\ref{fig:model_updates_per_hour}. We design our system for fast model aggregation that can support much higher throughput of server model updates (Section~\ref{sec:stateful_persistent_aggregation}) than what typical \syncfl systems can achieve.

In the next sections, we describe the design of our production system and explain how it supports the four requirements above. 


\section{System Components}
\label{sec:system_components}
\begin{figure}[t]
     \centering
     \begin{minipage}[t]{\linewidth}
         \centering
         \includegraphics[width=\linewidth]{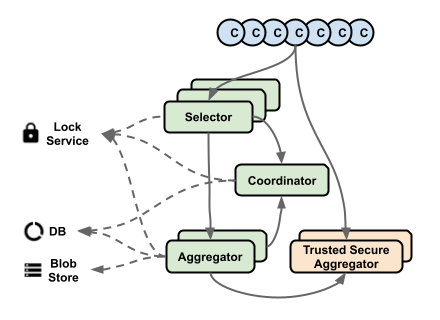}
     \end{minipage}
     \caption{\papaya high-level architecture.}
     \label{fig:syntem}
\end{figure}

The \papaya high-level design involves two applications: a server application that runs on a server in the data center, and the client application that runs on end-user devices. The server has three main components: Coordinator, Selector, and Aggregator. While the number of Selectors and Aggregators can scale elastically based on the workload demand, there is only one Coordinator; see Figure~\ref{fig:syntem}.

\papaya's system architecture is influenced by the Google FL stack (GFL) described in~\citet{google-fl}. We use the same names for the main components, and their functions are similar to those of GFL. However, their implementation and interactions are substantially different. GFL supports only \syncfl, whereas \papaya supports both SyncFL and AsyncFL. As a result, our design has fundamental differences in the protocol, execution, and scalability which enable it to achieve faster model convergence and straggler resilience; these differences are discussed further in Section \ref{sec:related}. First, we briefly describe the responsibilities of the main components and their interactions.

\textbf{Coordinator.} The Coordinator performs three main functions. First, it assigns FL tasks to Aggregators, as discussed in Section~\ref{sec:stateful_persistent_aggregation}. Second, the Coordinator assigns clients to FL tasks, as described in Section~\ref{sec:client_independence}. Finally, it provides centralized coordination and ensures that tasks progress in the face of Aggregator failures. 

\textbf{Selector. } The Selector is the only component that directly communicates with clients.  When necessary, it forwards client requests to other components. The Selector has two main responsibilities. For client selection, it advertises available tasks to clients, and summarizes current client availability for the Coordinator, as described in Section~\ref{sec:sustained_high_concurrency}. For client participation, the Selector routes client requests to the corresponding Aggregator, as described in Section~\ref{sec:stateful_persistent_aggregation}. 

\textbf{Aggregator. } Every task is assigned to a single Aggregator for the duration of the task (apart from failures and network partitions), as described in \cref{sec:stateful_persistent_aggregation}. The Aggregator has three main responsibilities. First, it aggregates client model updates to produce new versions of the server model. Second, it drives participating clients to run the client execution protocol, as described in \cref{sec:client_independence}. Finally, it tracks whether or not a task needs more clients and reports this to the Coordinator, as discussed in~\cref{sec:sustained_high_concurrency}. 

\textbf{Client Runtime.} The client runs on end-user devices and monitors training eligibility criteria such as whether or not the device is idle. It also tracks prior participation history to enable fair and unbiased client selection. If a client is eligible for training, the client checks in with the server to execute the FL client protocol as described in \cref{sec:client_independence}.


\section{Secure Aggregation}
\label{sec:secagg}
\label{sec:async_secagg}
\begin{figure}[t]
     \centering
         \includegraphics[width=0.8\linewidth]{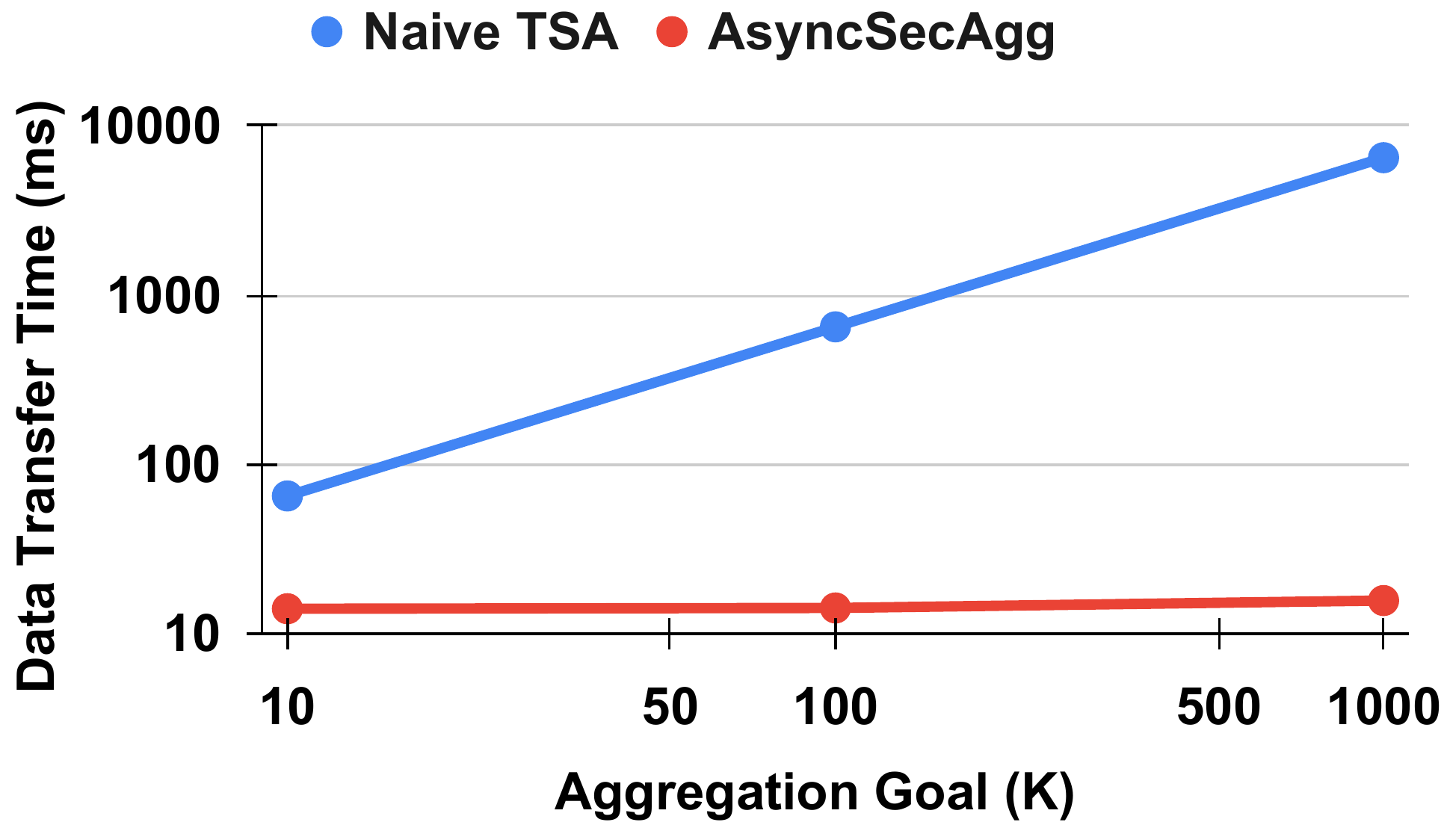}
     \caption{Data transfer time versus aggregation goal to transfer data across boundary into TEE for a 20MB model. We ran a benchmark to obtain the data transfer time for $K=1$ and use that to extrapolate other points in this figure, as naive TEE's data transmission is linear in $K$. Transferring the full model from each client to the TEE (Naive TSA) would take around 6500 milliseconds just for data transfer (when aggregation goal is 1000). In AsyncSecAgg each client only sends a 16-byte seed to the TEE, independent of model size.
     In this figure, the trusted hardware resides in the same machine as server. The latency is potentially greater if the trusted hardware is through a cloud provider. } 
     \label{fig:naive_secagg_data_transfer}
\end{figure}

In this section, we summarize our SecAgg mechanism to enable \asyncfl. In an honest-but-curious threat model, SecAgg allows the server to compute aggregated client updates without observing individual client updates. There are two main approaches for implementing SecAgg: using Secure Multiparty Computation (SMPC) or a Trusted Execution Environment (TEE). 

    Existing SMPC-based SecAgg approaches \cite{secagg, bell2020secure, so2021turbo} hinder asynchronous training, as they require cohort formulation and inter-client communication in each round.\footnote{A concurrent work ~\cite{so2021secure} describes an SMPC method that may overcome some of these issues. This approach could be an alternative to the TEE-based approach described here.} Meanwhile, \asyncfl does not have a discrete notion of rounds; clients join and finish training asynchronously. 

On the other hand, naive TEE aggregation is unscalable. Asymptotically, this approach transmits $O(K\cdot m)$ data across the host-TEE boundary, where $K$ is the aggregation goal and $m$ is the model size. Transferring data across the TEE boundary is time-consuming (\cref{fig:naive_secagg_data_transfer}): taking nearly 650 milliseconds for 100 clients ($K=100$), each with a 20MB model. This data transfer time increases with aggregation goal. Trusted hardware trades performance for security guarantees.


Motivated by these challenges, we propose an \emph{Asynchronous SecAgg} mechanism, relying on a TEE and an attestation mechanism; ensuring the Trusted Secure Aggregator (TSA) has not been tampered with. In this approach, \textit{random masking} relies on an additive one-time-pad to protect client updates and utilizes the TSA to generate aggregated random masks, unmasking aggregated client updates. The overall mechanism depends on a secure virtual channel established between each client and the TSA using the Diffie-Hellman key exchange protocol \citep{merkle1978secure}. Then, the mechanism leverages the TSA's ability to regenerate a random \emph{unmask} based on the clients' secret received by the TSA over the secure channel.

Asynchronous SecAgg empowers client independence and fast incremental aggregation. The protocol consists of the following steps: (1) A participating client establishes a secure virtual channel with the TSA and validates the secure aggregation configuration and integrity of the TSA; (2) The client shares the masked model update with a corresponding Aggregator and the random seed used to generate the mask with the TSA; (3) The aggregator incrementally aggregates masked model updates; (4) The aggregator requests the TSA to generate the unmasking vector once the configured aggregation goal is reached; (5) The aggregator unmasks the aggregated model updates using the unmasking vector and creates a new server model.

The random seed, usually 16 bytes shared between each client and the TSA, allows the two parties to share an as-large-as-the-model mask at a constant cost. Asymptotically, this approach only transmits $O(K+m)$ data across the boundary of the TSA. Appendix~\ref{sec:protocol_design_and_security_proof} presents more details about our secure aggregation protocol, including a security proof.

\section{System Design}
\label{sec:system_design}



In this section, we describe the system requirements to run \asyncfl at scale and the design choices we made to fulfill these requirements. We focus on the three most important requirements for \asyncfl outside of Asynchronous SecAgg (discussed in Section~\ref{sec:secagg}). For completeness, other requirements for running \asyncfl are described in  Appendix~\ref{appendix:system_design}. 

There are three main requirements. First, \asyncfl relies on clients training asynchronously. Hence, the client protocol must not introduce any dependence between clients. Second, \asyncfl can support higher client utilization than \syncfl. To enable this, our system must perform fast client replacement. Third, \asyncfl takes many more server model steps than \syncfl per unit time. Hence, our system must support fast model aggregation.

\subsection{Client Independence}
\label{sec:client_independence}
To enable asynchronous training, \papaya's client protocol deliberately avoids any inter-client dependency. Moreover, transient client failures do not cause clients to dropout because the client protocol is based on virtual sessions instead of persistent connections. At a high level, the protocol can be split into two phases: selection and participation. To explain the selection process, we first define \textit{client demand} for a task as the difference between the target concurrency and the number of users already participating in the training of the task. 

\textbf{Selection.} For a client, the goal of the selection phase is to find a task with positive client demand. Thus, a client can complete the selection phase with either \textit{acceptance} (client is accepted for participation) or \textit{rejection} (client will try to participate at another time). 

\textbf{Participation.} Once a client is accepted, the goal of the participation phase is for a client to share a trained model with the server. Participation consists of four stages. 1.~A client first \textit{downloads} model parameters, model code and configuration from a content delivery network. 2.~Next, the client \text{trains} the downloaded model on its local data. 3.~Once the training finishes, the client \textit{reports} its status to the server. The server shares an upload configuration with the client and, if enabled, the SecAgg configuration. 4.~In the final stage, the client \textit{uploads} the model in chunks, potentially after masking the model if SecAgg is enabled. All stages happen within a virtual session established during selection.

\begin{figure}[t]
     \centering
         \centering
         \includegraphics[width=0.8\linewidth]{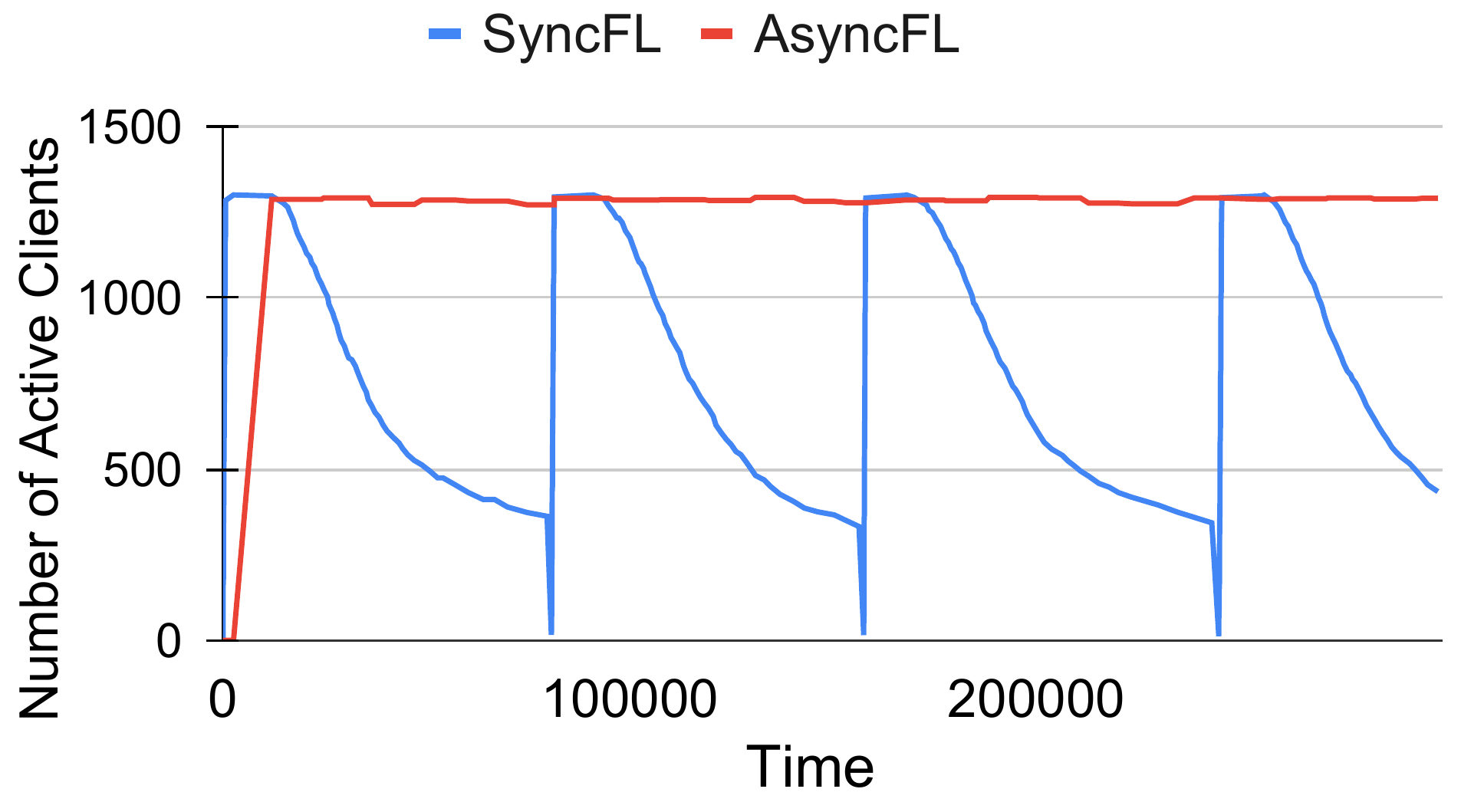}
     \caption{AsyncFL achieves high client utilization while SyncFL client utilization fluctuates. SyncFL proceeds in \emph{rounds}. The number of active clients (client utilization) increases at the beginning of a round as clients join the cohort, and it falls gradually towards the end of the round due to stragglers. In AsyncFL, the number of active clients stays relatively constant over time; as clients finish training and upload their results, other clients take their place. Both configurations in the figure have max concurrency of 1300. SyncFL uses 30\% over-selection.}
     \label{fig:sync_async_concurrency}
\end{figure}

\subsection{High Client Utilization}
\label{sec:sustained_high_concurrency}
\asyncfl is capable of higher client utilization compared to \syncfl. This is mainly because in \syncfl the number of active clients increases at the beginning of a round as clients join the cohort, and it falls gradually towards the end of the round as the server waits for all clients to finish training (\cref{fig:sync_async_concurrency}). On the contrary, in \asyncfl there is no cohort formation; as soon as one client completes training or fails, a new one is selected. Thus \asyncfl achieves high utilization throughout training. We show in~\cref{fig:sync_async_concurrency} that utilization in our \asyncfl implementation is close to 100\% throughout training. 
To realize high utilization, an \asyncfl system needs to replace completed and failed clients quickly. Achieving high utilization is especially challenging in a multi-tenant FL system, where multiple FL tasks are running in parallel, and a single client may be compatible with many tasks. 


We now describe the client assignment process which is responsible for maintaining high utilization. There are three important steps to assigning clients to tasks: tracking client demand for each task, tracking task eligibility for each client, and performing the actual assignment. 

\textbf{Tracking client demand for each task.} First, each Aggregator tracks client demand for the tasks that are assigned to it. When a client finishes training or fails, the Aggregator increases client demand for the associated task. Next, the Coordinator pools together information from all Aggregators into a consolidated view of client demand for every task in the system. Note that the Coordinator must explicitly account for clients that have been assigned to a task, but have not yet confirmed the assignment. 

\textbf{Tracking task eligibility for each client.} For each available client, the Coordinator constructs a list of eligible tasks. A task is eligible if the client is compatible with its requirements (e.g., can train the model of the task), and if the task has positive client demand.

\textbf{Task assignment.} Once an eligible task list is constructed for a client, the Coordinator randomly assigns the client to an eligible task. Concretely, the Coordinator instructs Selectors to forward the client to the Aggregator responsible for the task.

\subsection{Fast Model Aggregation}
\label{sec:stateful_persistent_aggregation}

\begin{figure}[t]
     \centering
         \includegraphics[width=0.8\linewidth]{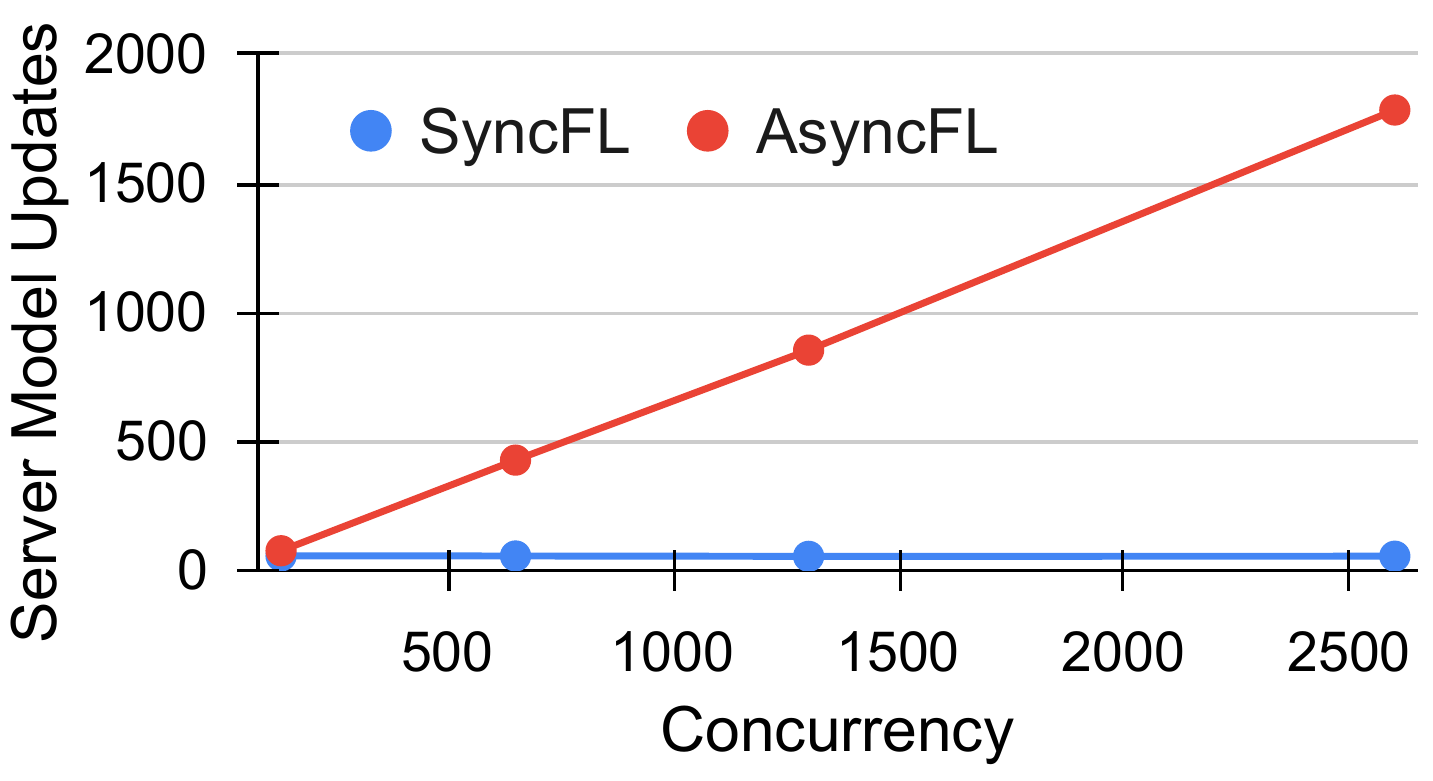}
     \caption{ Server Model Updates per hour with concurrency. At a concurrency of $2{,}300$, \asyncfl generates roughly 30$\times$ more server model updates per hour. The aggregation goal for \asyncfl is fixed at 100.}
     \label{fig:model_updates_per_hour}
\end{figure}

As shown in Figure~\ref{fig:model_updates_per_hour}, \asyncfl generates server model updates up to 30$\times$ more frequently than \syncfl.  Thus, fast model aggregation in a scalable \asyncfl system is critical. In this section, we describe how \papaya efficiently aggregates client updates. 

\textbf{Persistent Aggregator.} In our system, Aggregators are persistent and stateful because creating a new Aggregator for each task incurs a substantial overhead. Therefore, the Coordinator moves tasks between Aggregators only when it detects failed or overloaded Aggregators. The Coordinator evenly distributes tasks among available Aggregators using the estimated workload of a task. The Coordinator estimates this workload using the task concurrency and model size.


\textbf{Parallel Model Aggregation.} Once a client completes training, it uploads the trained serialized model update to the server. This update is then pushed into an in-memory queue on the Aggregator. A different thread drains the queue by de-serializing the updates into trainable parameters and aggregating them. To speed up this aggregation, we parallelize the aggregation process across available cores. To reduce lock contention, the ID of the thread performing intermediate aggregation is hashed to choose one of the intermediate aggregates. Once the cumulative number of aggregated model updates reaches the aggregation goal, the final aggregation is performed and a new server model is generated. Note that the aggregation goal in \syncfl is typically $1.3 \times$ concurrency (30\% over-selection), while in \asyncfl it is independent of concurrency.

\section{Evaluation}
\label{sec:evaluation}
\begin{figure*}[t]
     \centering
     \begin{minipage}[t]{\textwidth}
         \includegraphics[width=0.33\textwidth]{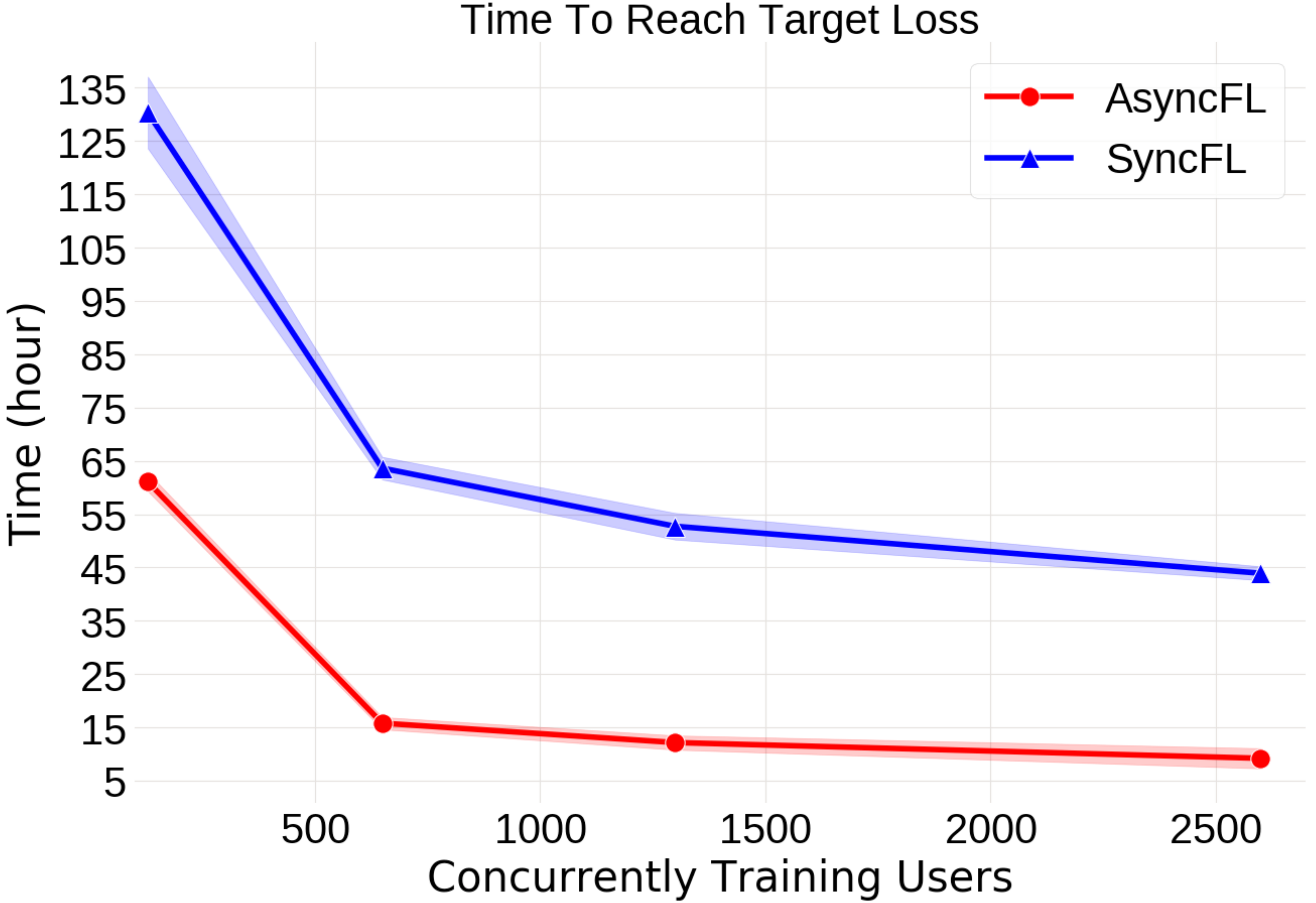}
         \includegraphics[width=0.33\textwidth]{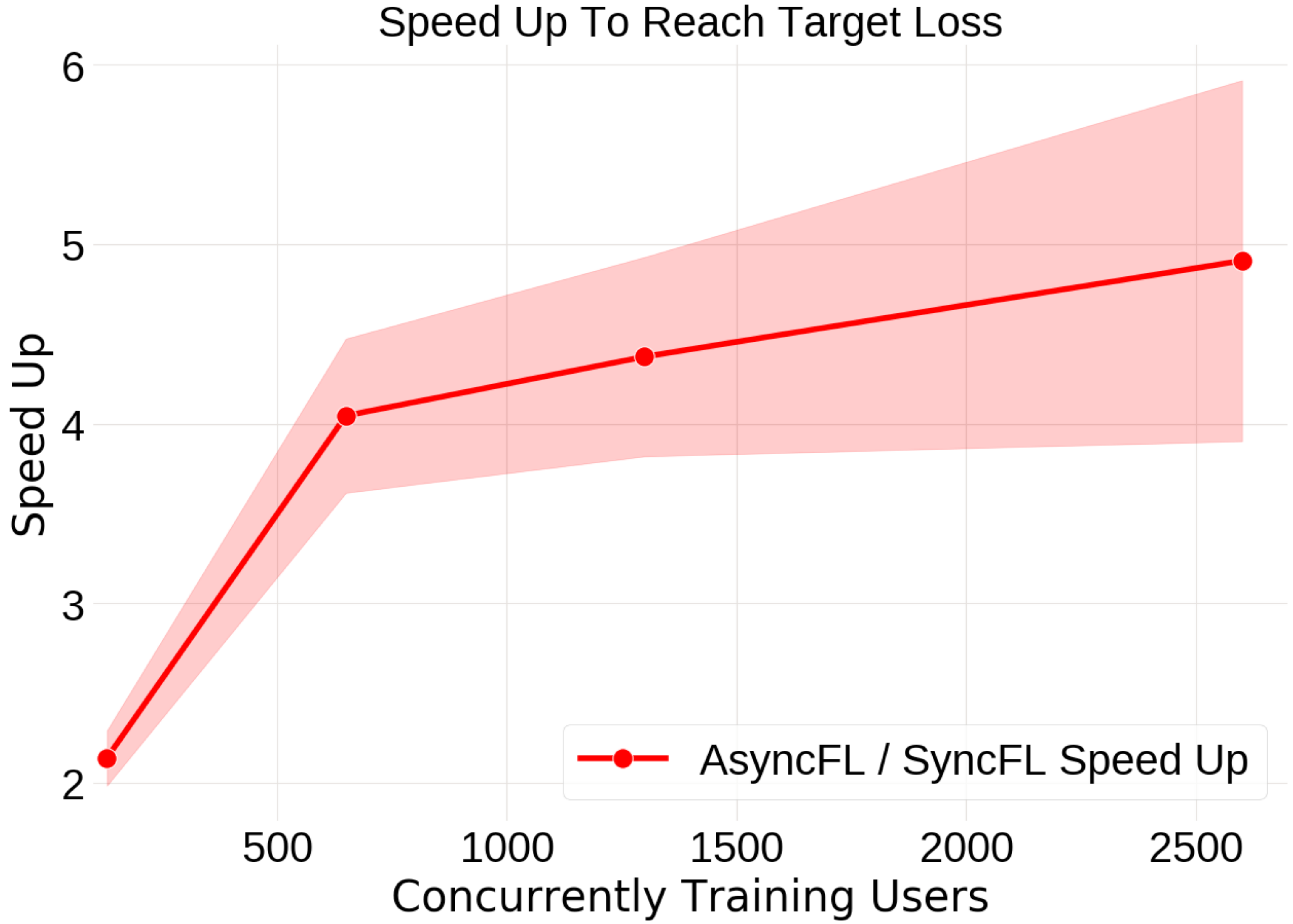}
         \includegraphics[width=0.33\textwidth]{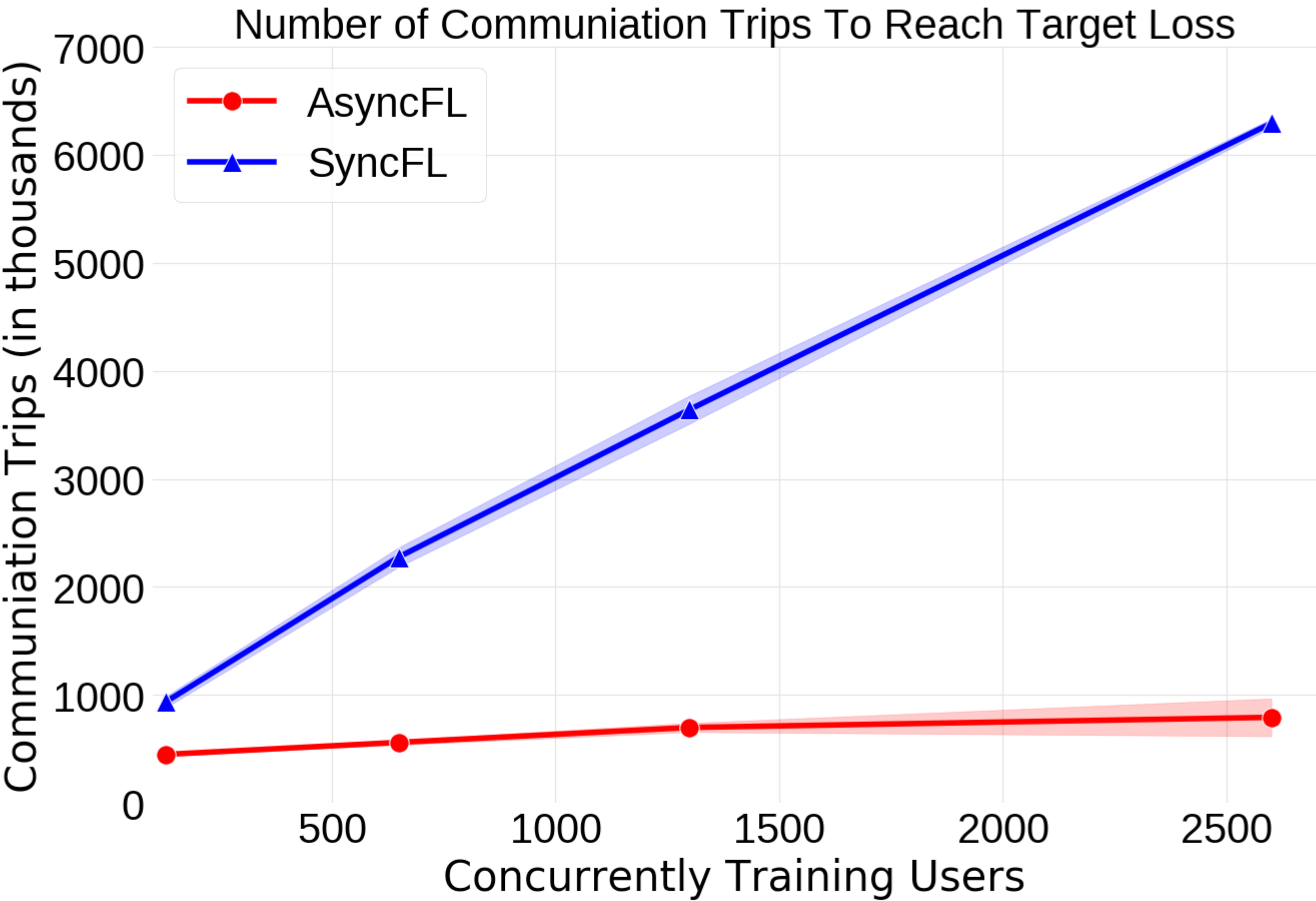}
     \end{minipage}
     \caption{(left) Number of hours to reach a target loss. For AsyncFL, a server update is produced every 100 client updates, $K$ = 100. For SyncFL, we use 30\% over-selection as in \cite{google-fl} to mitigate stragglers. (middle) Speed up of AsyncFL relative to SyncFL. AsyncFL is 5x faster than SyncFL. (right) As concurrency increases, AsyncFL outperforms SyncFL by increasingly larger amounts. }
    \label{fig:convergence}
    \label{fig:stragglers}
\end{figure*}

In this section, we present evaluation results for AsyncFL. We first compare the convergence speed, scalability, and communication efficiency of AsyncFL with SyncFL. Next, we analyze the source of AsyncFL's speed up. We then show that SyncFL can be either straggler resilient (with over-selection) or be unbiased (without over-selection), but not both simultaneously. In contrast, AsyncFL is straggler resilient without introducing bias.

\subsection{Experimental Setup}
To study the performance of AsyncFL, we train an LSTM-based language model \cite{char_aware_lm}, a common FL application \cite{gboard-fl}, on a population of nearly 100 million Android phones. We repeat each experiment 3 times, each at the same time of the day, and report the average. AsyncFL and SyncFL are run at the same time, so they have access to the same client population.

Following the requirements in~\citet{gboard-fl}, a client device can participate in FL training only when idle, charging, and on an unmetered network. Similar to~\citet{google-fl}, a timeout is imposed to limit the client training time; we set the timeout to 4 minutes. The distribution of client execution times is analyzed in Section~\ref{sec:sampling_bias}. 

For both SyncFL and AsyncFL, we use SGD on the client and FedAdam \cite{adaptive-fl-optimization} on the server. For the server optimizer, we use Adam's default learning rate and tune the first-moment parameter in simulation. We run hyperparameter sweeps in simulation, using a representative dataset, for the client optimizer to find the best client learning rate. Each client runs one local epoch of training with batch size $B = 32$. We partition each client's data into train, test, and validation sets randomly. 

Our system has two configuration parameters. First, for both SyncFL and AsyncFL tasks, concurrency specifies the maximum number of concurrently participating devices. Second, for AsyncFL tasks, $K$ is the aggregation goal, controlling the size and frequency of server update. The server produces a new model every $K$ client model updates. In our experience, we find that choosing $K$  to be 10-30\% of concurrency works well in practice. Finally, unless otherwise stated, we use 30\% over-selection with SyncFL to alleviate the impact stragglers, as proposed in \citet{google-fl}. 


\subsection{Results on Convergence Time and Scalability}
\label{sec:scalability}

\begin{figure}[t]
     \centering
     \begin{minipage}[t]{\linewidth}
        \includegraphics[width=\linewidth]{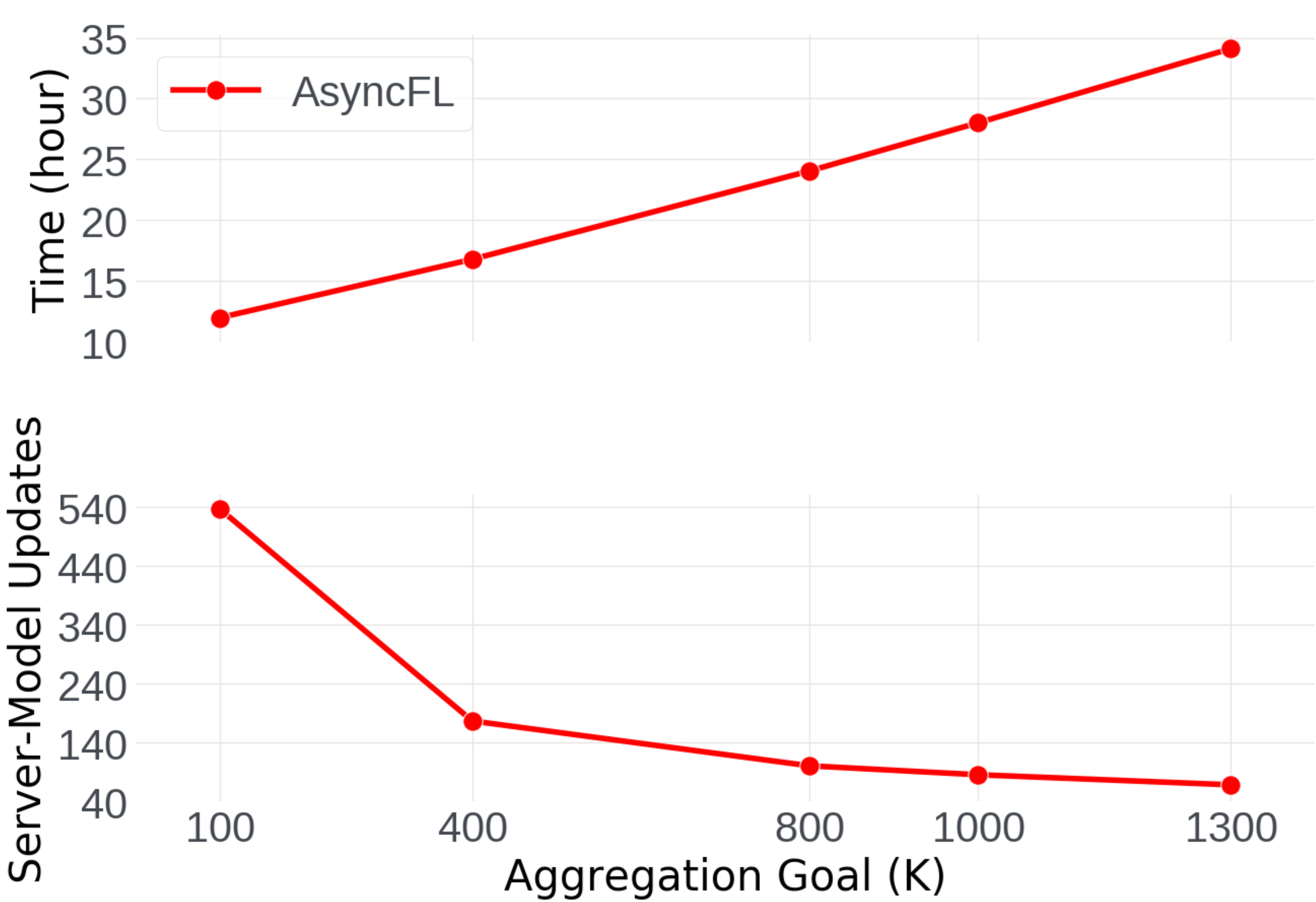}
     \end{minipage}
     \caption{(top) Number of hours to reach a target perplexity of 60 with concurrency $= 1{,}300$ and varying values of $K$. (bottom) Server Model Updates per hour at concurrency $= 1{,}300$ and varying values of aggregation goal $K$.}
     \label{fig:different_k}
\end{figure}

To begin, we evaluate the training time performance and scalability of AsyncFL and SyncFL. We measure the convergence time as the wall-clock training time to reach a target loss. To measure scalability, we compare AsyncFL and SyncFL in terms of their speedup and the number of communication trips with increasing concurrency. \Cref{fig:convergence} shows (left) the time taken by the two algorithms to reach a target loss for varying levels of concurrency, (middle) the speedup of AsyncFL over SyncFL, and (right) the number of communication trips to reach a target loss. As \Cref{fig:convergence} (left and middle) illustrates, the speedup gap widens as concurrency increases, from $2\times$ to $5\times$.
Furthermore, the SyncFL communication efficiency worsens. The overall communication efficiency gain of \asyncfl increases from $2\times$ to $8\times$ as concurrency increases. The evaluation results demonstrate that \asyncfl handles the system heterogeneity and scalability challenges more effectively than SyncFL. In the following sections, we unravel why our AsyncFL system is more suitable for scaling FL training to hundreds of millions of clients.

\subsection{Analysis of Server-Model Step Frequency}
\label{sec:server_step}
To understand the performance of our AsyncFL implementation in detail, we study how the aggregation goal impacts convergence. The aggregation goal determines how many client updates contribute to each server model update, and for a fixed concurrency, it also affects the frequency of server model updates. We fix concurrency to be 1300 and vary aggregation goal ($K$) from 100 to 1300. \Cref{fig:different_k} (top) depicts the time for each configuration to reach a target loss, while \Cref{fig:different_k} (bottom) describes the server-model step frequency per hour. As $K$ increases, the batch size increases, and the server takes less frequent model steps.  Thus, the larger the $K$ is, the slower the convergence time. 
It is natural to ask if convergence time could be further reduced for $K$ smaller than 100.
However, \citet{fedbuff} found that moderate values of $K$ can lead to more stable convergence.
Moreover, the frequency of server updates is limited by the system's write bandwidth. Thus, we cannot create a new server model too often. We leave improvements to overcome write bandwidth limitations as future work.


\subsection{Analysis of Sampling Bias from Over-Selection}
\label{sec:sampling_bias}

\begin{table}
\centering
\caption{Test perplexity (lower is better) after 1 million client updates. Perplexity is a measure of language model accuracy. We partition clients into percentiles, based on the training data volume. All signifies all clients; 75\% and 99\% represent clients with data volume in the 75th and 99th percentiles, respectively. SyncFL w/o OS denotes \syncflnoos and SyncFL w/ OS denotes \syncflos}
\begin{tabular}{lrrrrrr} 
	\toprule
	Method                     & All & 75\% & 99\% & Time (hour) \\ 
	\midrule
	SyncFL w/o OS & 68.38                   & 66.64                                     & 47.82                                  & 130.60                \\
	SyncFL with OS & 72.97                   & 73.10                                     & 73.24                                  & 18.63                 \\
	AsyncFL          & 57.32                   & 55.71                                     & 38.51                                  & 18.28                 \\
	\bottomrule
\end{tabular}
\label{table:bias}
\end{table}


\begin{figure}[t]
     \centering
     \begin{minipage}[t]{\linewidth}
         \centering
         \includegraphics[width=\linewidth]{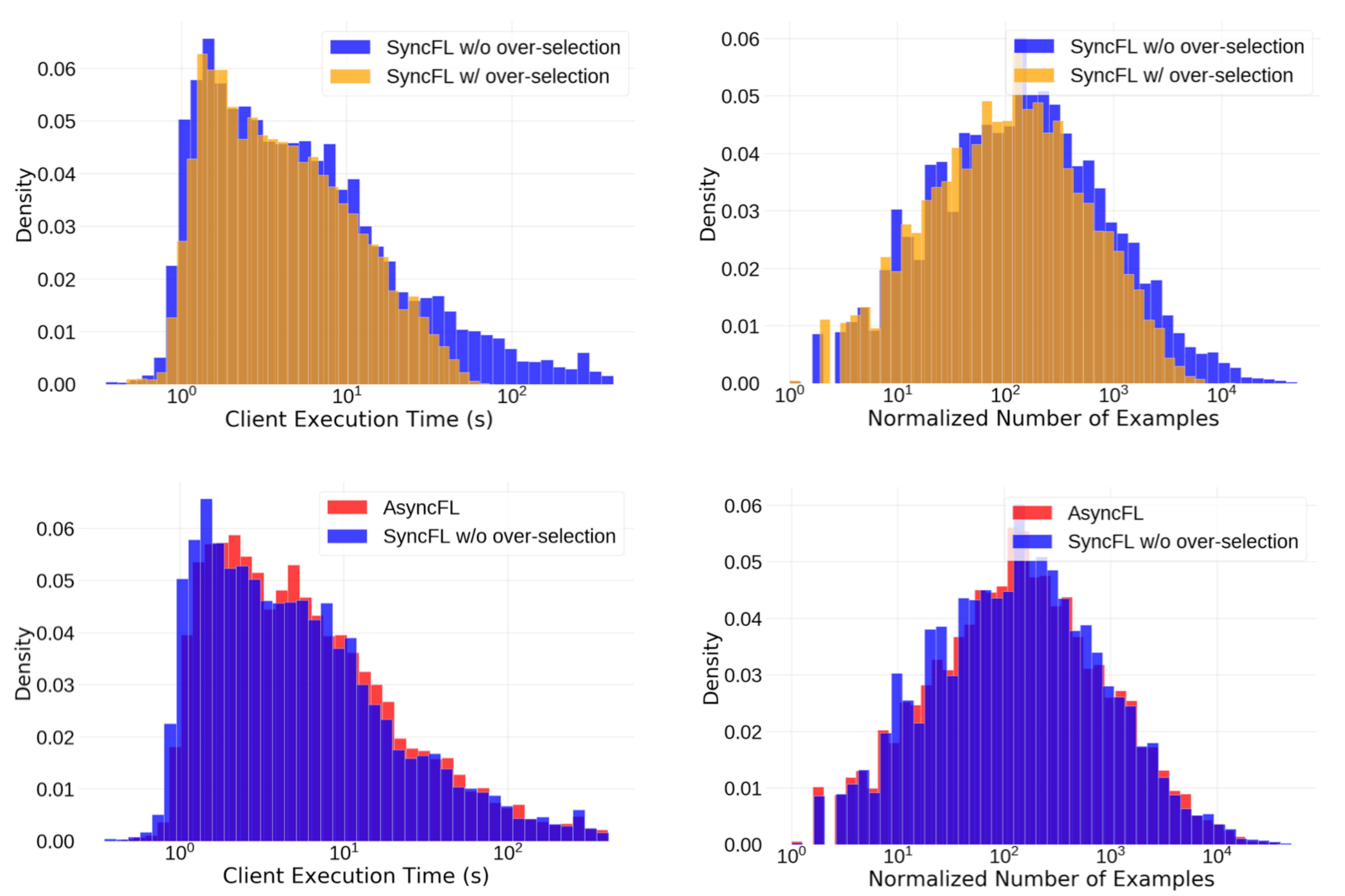}
     \end{minipage}
     \caption{(top) Participating client execution time and normalized number of examples for \syncflos and \syncflnoos. (bottom) Histogram of number of training examples for participating clients for SyncFL and AsyncFL. In the right figure, SyncFL with 30\% selection drops the slowest clients. The slowest clients are often the ones with many training examples, as illustrated in the right figure.}
     \label{fig:over_selection}
\end{figure}

To compare the effectiveness of over-selection and asynchronous training in combating stragglers, we examine the distribution of participating clients, their execution time, and the number of training examples. \Cref{fig:over_selection} illustrates the discrepancy between the client execution time distribution of SyncFL with and without over-selection. Since over-selection discards updates from some clients, the distribution of SyncFL without over-selection should be considered representative of the entire client population.  \Cref{fig:over_selection} (top-left) shows that over-selection drops slow clients, as desired. However, as illustrated in \Cref{fig:over_selection} (top-right), the slowest clients often have more training examples. 

To rigorously assess the difference, we perform a two-sample Kolmogorov-Smirnov test~\cite{kstest} to measure the goodness of fit between AsyncFL, SyncFL with over-selection, and the ground truth distribution (SyncFL without over-selection). We find that the D-statistic, representing the absolute max distance between the cumulative distribution functions of the two samples, for \asyncfl and the ground truth is $8.8\times 10^{-4}$ ($p$-value = 0.98). In comparison, the D-statistic for \syncflos and the ground truth is $6.6\times 10^{-2}$ ($p$-value = 0.0). This result shows that \asyncfl and the ground truth have similar distributions while \syncflos does not. 
Thus, over-selection introduces sampling bias while AsyncFL does not. The sampling probability is conditioned on the client's device speed or the number of training examples. Next, we show that sampling bias hurts model performance, especially for clients with more training examples. 

\Cref{table:bias} reports the model quality in test perplexity for all clients and those with data volume in the 75\% and 99\% percentile. Perplexity is a measure of language model accuracy (lower is better). The sampling bias from over-selection in SyncFL causes a 6\% drop in model quality overall and a 50\% drop in model quality for clients with more examples. Although \syncflnoos is unbiased, it is also 10$\times$ slower. On the other hand, AsyncFL combines fast training with high model quality and no sampling bias. Meanwhile, \syncflos has to choose between sampling bias or straggler resilence. In summary, AsyncFL is a more desirable method to address the impact of stragglers. 

\begin{figure}[t]
     \centering
     \begin{minipage}[t]{\linewidth}
        \includegraphics[width=\linewidth]{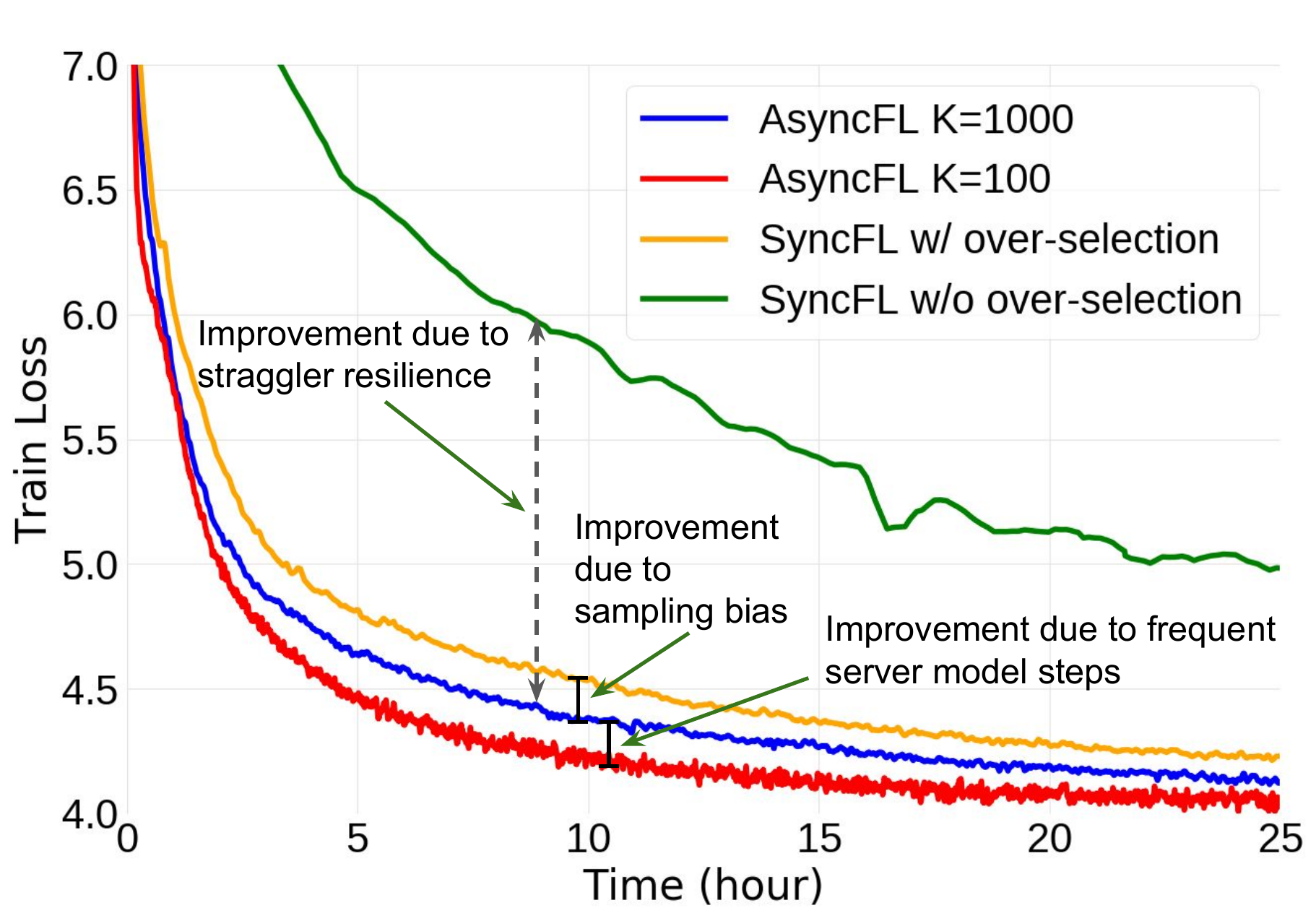}
     \end{minipage}
     \caption{Training curves for different FL configuration at aggregation goal $= 1{,}000$. We set concurrency $= 1{,}300$ for \asyncfl, \syncflos. For \syncflnoos, we set concurrency equal to aggregation goal.}
     \label{fig:source_of_speed_up}
\end{figure}

\begin{figure}[t]
     \centering
     \begin{minipage}[t]{\linewidth}
        \includegraphics[width=\linewidth]{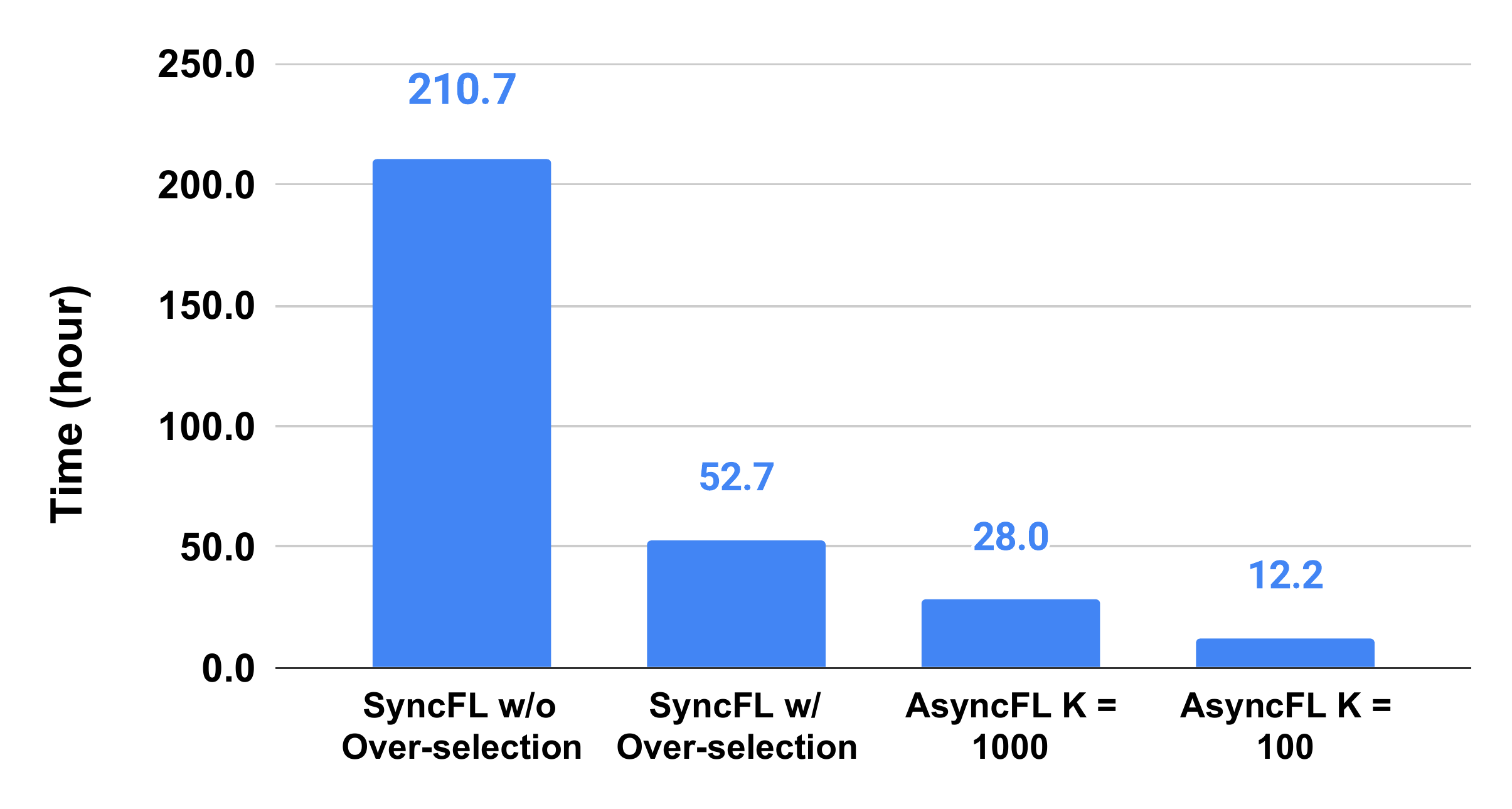}
     \end{minipage}
     \caption{Number of hours to reach a target loss for different FL design configurations. }
     \label{fig:final_speedup_comparison}
\end{figure}

\subsection{Understanding AsyncFL Advantages}
\label{sec:source_of_speed_up}
The previous sections showed that \asyncfl has two main advantages over \syncfl: better scalability because of more frequent model steps and straggler resilience without adding sampling bias. 
To quantify the benefits from these two properties, we present the training curves for \asyncfl alongside the current state-of-art \syncfl. We remove the frequent update advantage of \asyncfl by increasing the aggregation goal for \asyncfl to be the same as \syncfl. 

\Cref{fig:source_of_speed_up} depicts production training curves for the best synchronous setup, \syncflos (orange), and two \asyncfl configurations: aggregation goal $K$ = 100 (red) and $K$ = 1000 (blue). All three use concurrency 1,300. Note that overall, \asyncfl with $K$ = 100 is 4.3$\times$ faster than \syncflos, as shown in \Cref{fig:final_speedup_comparison}. We find that about half of this speedup comes from using smaller $K$ and the rest from avoiding sampling bias (i.e., using \asyncfl rather than \syncflos).

To read Figure~\ref{fig:source_of_speed_up}, start with \asyncfl with $K$ = 100 (red), which is the best configuration since it takes more frequent server model steps and is resilient to stragglers. Next, see \asyncfl with $K$ = 1000 (blue), which is straggler resilient but takes less frequent model steps. Finally, move to \syncflos (orange), which adds sampling bias. 

The figure also shows \syncflnoos (green) for reference, using concurrency 1000. The large gap between this configuration and \asyncfl with $K$ = 1000 is attributable to stragglers.

It is instructive to compare the training loss at a fixed point, e.g., at the 10-hour mark. By minimizing sampling bias, \asyncfl with $K$ = 1000 reduces training loss by 3.4\% compared to \syncflos. Taking more frequent server-model steps ($K$ = 100) in \asyncfl decreases training loss by an additional 3.5\%.  



\section{Related Work}
\label{sec:related}
The \papaya system described in this paper is inspired by the GFL system \cite{google-fl}. Another FL system is described by Apple (AFL) in \cite{apple-fl}. We focus on comparison with GFL and AFL given the similarity of production scale. At a high level, both GFL and AFL only implement SyncFL, while \papaya implements both \syncfl and \asyncfl. Diving deeper we find similarities and differences in how clients are selected for participation, client availability and participation outcome impact on model training progress, model update aggregation and privacy mechanisms. \papaya actively (through the Coordinator) selects available clients for participation at any point in time based on demand by active tasks (driven by desired task concurrency), unlike GFL which actively selects clients before the rounds starts, and AFL which uses passive probabilistic selection. \papaya enables incremental progress by making participating clients independent and replaceable whenever they complete or drop out, unlike GFL where no client can join after a round has started, potentially leading to failed rounds, and similar to AFL where clients can contribute as long as the model version is the same. \papaya moves tasks between long living Aggregators only when failure or load imbalance is detected to minimize client progress loss and reduce placement overhead, unlike GFL where tasks are dynamically placed to ephemeral Aggregators every round and AFL where aggregation is performed by an offline service. \papaya implements Asynchronous SecAgg based on TEEs, whereas GFL uses SMPC-based Synchronous SecAgg and AFL does not report using SecAgg.    

Another line of related work is the FL software tool kits, offered by other technology companies. Notable among these are Clara~\cite{clara}, IBM-FL~\cite{ibm-fl}, OpenFL ~\cite{openfl-full} and FATE~\cite{fate}. While related to the \papaya system described in this paper, these software tools are distinct from production FL systems training across hundreds of millions of devices, which is the focus of this paper.


\section{Conclusions}
\label{sec:conclusions}
We presented our design of a production asynchronous FL system for training at scale. Designing for a production FL system, \papaya, we find that AsyncFL is faster, more straggler resilient, and provides better model quality than SyncFL. \papaya is flexible and supports both synchronous and asynchronous FL. Empirically, we demonstrated that in high concurrency settings, asynchronous FL achieves 5$\times$ faster speed up and conserves nearly 8$\times$ more resources than synchronous FL. Finally, \papaya can be extended with features to enable differential privacy, which we leave as future work.

\section*{Acknowledgements}
We thank Ilya Mironov and Rachad Alao for meaningful discussions and their valuable support which significantly improved the quality of this paper. 




\bibliography{reference}
\bibliographystyle{mlsys2022}

\appendix
\clearpage
\section*{Supplementary Material}
\label{sec:appendix}

\section{Cryptographic Primitives}
\label{sec:crypto_primitives}

\subsection{Diffie–Hellman Key Exchange Protocol}

Diffie–Hellman key exchange protocol allows two parties to securely agree on a randomly-generated shared secret via an untrusted communication channel. Viewed in the server-client setting, the protocol consists of an initial message from one party (server) and a completing message as a response from the other one (client). The server can prepare the initial messages in advance, without knowing the identities of the clients. The client can solely determine the shared secret once it receives the initial message. The client  needs to interact with the server only once to finish the protocol by sending the completing message.

\subsection{Additive One-time pad (OTP)}

There are many existing additive homomorphic encryption schemes such as the works in \cite{elgamal1985public,GM82, paillier1999public, cohen1985robust}. The complexity of the decryption algorithms in these protocols is usually linear in the ciphertext size and is independent of the number of additions. However, these schemes often operate on a large finite group whose elements can be as large as 1024--3072 bits. Such requirement inflates the ciphertext size even if the plaintext space is much smaller (e.g., 32 bit integers). Such blow-up makes these schemes less desirable when ciphertexts are transmitted via network and traffic is at a premium, for example, on mobile devices. 

A PRNG-generated additive one-time pad (OTP) is a good alternative to avoid the expansion of the ciphertext. The protocol is summarized in \cref{fig:additive_one_time_pad}.

The additively homomorphic encryption scheme in \cref{fig:additive_one_time_pad} can operate over any finite Abelian group (e.g., $\mathbb{Z}_{2^{32}}$). Therefore the ciphertext can be in the same space as plaintext. The downside is that the complexity of the decryption algorithm scales up linearly with number of additions performed, in contrast to a constant in other encryption schemes. We argue that trading in decryption workload for a more compact ciphertext is an acceptable trade-off in settings with mobile devices if the decryption is performed server-side for the following reasons:
\begin{enumerate}
  \item Mobile devices are often restricted in both computation power and communication bandwidth. An additive OTP is more efficient in both computation and bandwidth cost, compared to the group operations needed in other encryption schemes. 
  \item The server usually has much more computation resources to perform the relatively more expensive decryption. Furthermore, hardware acceleration optimizations are often available server-side, reducing the costs of the decryption algorithm.
\end{enumerate}

\begin{figure}
  \centering
\fbox{   \begin{minipage}{0.95\linewidth}
  {\bf Public parameters:} finite Abelian group $\mathbb{G}$
  \begin{itemize}
  \item {\bf $\mathbf{Enc_{k}(v)}$:} To encrypt a vector $v\in \mathbb{G}^{\ell}$, a cryptographically secure PRNG is used to generate a vector $m\leftarrow \mathbf{PRNG}(k)$ where $m \in \mathbb{G}^{\ell}$. The ciphertext $c$ is defined as the element-wise sum $v+m$.
  
  \item {\bf Addition:} Two ciphertexts $c_1$ and $c_2$ can be added together element-wise.
  
  \item {\bf Decryption:} An (aggregated) ciphertext $c\coloneqq \sum \mathbf{Enc}_{k_i}(v_i)$ can be decrypted as $\sum v_i = c - \sum \mathbf{PRNG}(k_i)$.
  \end{itemize}
  \end{minipage}
  }
    \caption{Additive one-time pads.}
  \label{fig:additive_one_time_pad}
\end{figure}

\section{Protocol Design and Security Proof}
\label{sec:protocol_design_and_security_proof}

In this section we will go over the detailed design of our protocol and formally prove its security. We adopt the same strategy from Cryptonite \cite{karl2020cryptonite}. A trusted party realized by trusted hardware (e.g., Intel SGX) will assist with the procedure and help make up for the dropped clients. With the assistance of the trusted hardware, clients no longer rely on each other to protect their own private inputs or mitigate the dropout of their peers. Without client interdependence, clients no longer need to communicate with each other via the server and no longer need to know about each others' identities. The absence of interdependency requirement among clients allows 
them to participate asynchronously, making our protocol compatible with FedBuff~\cite{fedbuff}.

\subsection{Problem Setup and Threat Model}
The thread model is composed of a server, a trusted third party and $n$ clients. The trusted third party and the clients can only communicate directly with the server. The clients can choose to participate in the protocol at any time, not necessarily in the beginning. Instead, they will check-in with the server when they become available. Clients may have limited availability. The availability of any two clients may have no overlap on the timeline.

Each client has a private $\ell$-element array of group elements of a finite group $G$, where $\ell$ and $G$ are public parameters. The trusted party's public key is available to all clients. The server and the trusted third party have no private inputs. The parties wish to collaborate and reveal the position-to-position aggregation result across at least $t$ clients' private array but any individuals' inputs should remain private. A malicious adversary may corrupt the server and number of clients.

The ideal functionality is summarized in \cref{fig:protocol_ideal_world}.
\begin{figure}
    \centering
\fbox{     \begin{minipage}{0.95\linewidth}
{\bf Public inputs:}
The group $G$, the vector length $\ell$ and threshold $t$.

{\bf Private inputs:}
Client $i$ has input $v_i \in G^{\ell}$. The server has no inputs.
\begin{enumerate}
\item The clients send their secret $v_i$ to the ideal functionality $\mathcal{F}$.
\item The ideal functionality $\mathcal{F}$ sends the list of clients $\mathcal{C}_0$ to the server.
\item The server chooses a subset of clients $\mathcal{C}_1 \subset \mathcal{C}_0$ and sends it back to the ideal functionality.
\item The ideal functionality $\mathcal{F}$ computes $\sum_{i \in \mathcal{C}_1} v_i$ and sends the sum to the server if $|\mathcal{C}_1| \ge t$ clients; otherwise, does nothing.
\end{enumerate}
    \end{minipage}
}
\caption{Ideal functionality  $\mathcal{F}$ for secure aggregation}
\label{fig:protocol_ideal_world}
\end{figure}

\subsection{Overview of Our Solution}
\label{sec:Protocol}

To avoid sending big chunks of data across the boundary of the secure enclave, we will aggregate random masks, instead of the actual data, inside the secure enclave. The high-level idea is to mask clients' private inputs with some additive masks, while the server will be responsible for aggregating the masked inputs and trusted party will be responsible for aggregating the masks. Note that a 128-bit seed is sufficient to represent a random mask. The amount of data transferred into the secure enclave for each client will be a constant, despite of amount of data to aggregate. Our protocol can be divided into three steps:

\begin{enumerate}
    \item New client checks in and validates the identity of the trusted party.
    \item Client sends masked input to the untrusted server and demasking information to the trusted party.
    \item The trusted party instructs the untrusted server how to demask the sum of all masked inputs.
\end{enumerate}

\subsection{Protocol Detail}

Our protocol is detailed in \cref{fig:protocol_real_world}. We use Diffie–Hellman key exchange protocol to establish private communication channels between the trusted party and the clients. 

\begin{figure}
    \centering
\fbox{     \begin{minipage}{0.95\linewidth}
    Client $i$ has input $v_i \in G^{\ell}$. The server and the trusted party have no inputs.
    \begin{enumerate}
        \item \label[step]{step:key_exchange_request} The trusted party runs $N (N > n)$ DH key exchange protocol instances and obtains $N$ DH key exchange initial messages. The trusted party then sends these initial messages with their indices and signatures to the server.
        \item  \label[step]{step:send_key_exchange_request} When the $i$'th client checks in with the server, the server sends the $i$'th initial message and the corresponding signature received from the trusted party to this client.
        \item  \label[step]{step:key_exchange_response} Upon receiving a DH key exchange initial message and the corresponding signature, client $i$ validates the signature and aborts if not valid. Otherwise, the client generates a DH key exchange completing message and obtains a secret $k_i$ that will be shared with the trusted party. 
        \item Client $i$ picks a random seed $s_i$ and uses it as the random seed to randomly generate $m_i \in G^{\ell}$, and sends $v_i+m_i$, $d_i \coloneqq \mathsf{Enc}_{k_i}(s_i)$ as well as DH key exchange completing message to the server. $\mathsf{Enc}$ employs standard techniques like MAC and sequential number to detect any tampered encryption.
        \item Upon receiving masked vector $v_i+m_i$, encryption $d_i$, and the DH key exchange completing message from client $i$, the server aggregates  $v_i+m_i$ to a running sum $\sum (v+m)$ and  sends the encryption $d_i$, the completing message, and the index of the corresponding initial message to the trusted party.
        \item \label[step]{step:update_seed} Upon receiving encryption $d_i$ and the completing message for the $i$'th initial message for DH-key exchange, the trusted party computes the shared secret $k_i$ and uses it to recover $s_i = \mathsf{Dec}_{k_i}(d_i)$. Then the trusted party re-generates $m_i$ with $s_i$ and aggregates it to a running sum $\sum m$. After that, the trusted party will not process any further completing messages to $i$'th initial message.
        \item \label[step]{step:generating_demasking_vector} The server can request the trusted party to generate the unmasking vector. Upon receiving such request, the trusted party sends the running sum $\sum m$ to the server only if at least completing messages of $t$ clients have been processed. The trusted party ignores any further messages from the server.
        \item \label[step]{step:output} Upon receiving $\sum m$, the server computes the sum of all private arrays by $\sum v =\sum (v+m) - \sum m$.
    \end{enumerate}
    \end{minipage}
    }
            \caption{Real world protocol for secure aggregation}
    \label{fig:protocol_real_world}
\end{figure}

\subsection{Security Proof}

\begin{figure*}[ht]
    \centering
\label{fig:simulator_illumination}
\tikzset{every picture/.style={line width=0.75pt}} 

\begin{tikzpicture}[x=0.75pt,y=0.75pt,yscale=-1,xscale=1]

\draw  [fill={rgb, 255:red, 155; green, 155; blue, 155 }  ,fill opacity=1 ][dash pattern={on 0.84pt off 2.51pt}] (280,137) -- (510,137) -- (510,383) -- (280,383) -- cycle ;
\draw  [fill={rgb, 255:red, 155; green, 155; blue, 155 }  ,fill opacity=1 ][dash pattern={on 0.84pt off 2.51pt}] (360,117) -- (430,117) -- (430,138) -- (360,138) -- cycle ;
\draw (395,127.5) node [anchor=center][inner sep=0.75pt]   [align=left] {Simulator};

\draw    (395,138) .. controls (368.35,152.64) and (348.6,129.04) .. (320,154) ;
\draw [shift={(320,154)}, rotate = 317.2] [color={rgb, 255:red, 0; green, 0; blue, 0 }  ][line width=0.75]    (10.93,-3.29) .. controls (6.95,-1.4) and (3.31,-0.3) .. (0,0) .. controls (3.31,0.3) and (6.95,1.4) .. (10.93,3.29)   ;

\draw  [fill={rgb, 255:red, 80; green, 227; blue, 194 }  ,fill opacity=1 ][dash pattern={on 4.5pt off 4.5pt}]  (290,154) -- (349,154) -- (349,190) -- (290,190) -- cycle  ;
\draw (319.5,172) node [anchor=center][inner sep=0.75pt]   [align=left] {\begin{minipage}[lt]{36.92pt}\setlength\topsep{0pt}
\begin{center}
Trusted\\ party
\end{center}
\end{minipage}};

\draw    (395,138) -- (395,154) ;
\draw [shift={(395,154)}, rotate = 270] [color={rgb, 255:red, 0; green, 0; blue, 0 }  ][line width=0.75]    (10.93,-3.29) .. controls (6.95,-1.4) and (3.31,-0.3) .. (0,0) .. controls (3.31,0.3) and (6.95,1.4) .. (10.93,3.29)   ;
\draw  [fill={rgb, 255:red, 80; green, 227; blue, 194 }  ,fill opacity=1 ][dash pattern={on 4.5pt off 4.5pt}]  (365,154) -- (425,154) -- (425,190) -- (365,190) -- cycle  ;
\draw (395,172) node [anchor=center][inner sep=0.75pt]   [align=left] {\begin{minipage}[lt]{35.04pt}\setlength\topsep{0pt}
\begin{center}
Honest\\client
\end{center}
\end{minipage}};

\draw    (395,138) .. controls (406.76,153.78) and (425.43,127.65) .. (470,154) ;
\draw [shift={(470,154)}, rotate = 210.87] [color={rgb, 255:red, 0; green, 0; blue, 0 }  ][line width=0.75]    (10.93,-3.29) .. controls (6.95,-1.4) and (3.31,-0.3) .. (0,0) .. controls (3.31,0.3) and (6.95,1.4) .. (10.93,3.29)   ;
\draw  [fill={rgb, 255:red, 80; green, 227; blue, 194 }  ,fill opacity=1 ][dash pattern={on 4.5pt off 4.5pt}]  (440,154) -- (500,154) -- (500,190) -- (440,190) -- cycle  ;
\draw (470,172) node [anchor=center][inner sep=0.75pt]   [align=left] {\begin{minipage}[lt]{35.04pt}\setlength\topsep{0pt}
\begin{center}
Honest\\client
\end{center}
\end{minipage}};

\draw  [dash pattern={on 0.84pt off 2.51pt}]  (319.5,190) -- (395,210) ;
\draw  [dash pattern={on 0.84pt off 2.51pt}]  (395,190) -- (395,210) ;
\draw  [dash pattern={on 0.84pt off 2.51pt}]  (470,190) -- (395,210) ;

\draw  [fill={rgb, 255:red, 74; green, 144; blue, 226 }  ,fill opacity=1 ]  (395, 240) circle [x radius= 55, y radius= 30]   ;
\draw (395,240) node [anchor=center][inner sep=0.75pt]   [align=left] {\begin{minipage}[lt]{50.92pt}\setlength\topsep{0pt}
\begin{center}
Real world\\protocol
\end{center}

\end{minipage}};
\draw  [fill={rgb, 255:red, 245; green, 166; blue, 35 }  ,fill opacity=1 ][dash pattern={on 4.5pt off 4.5pt}]  (290,290) -- (340,290) -- (340,316) -- (290,316) -- cycle  ;
\draw (315,303) node [anchor=center][inner sep=0.75pt]   [align=left] {\begin{minipage}[lt]{32.77pt}\setlength\topsep{0pt}
\begin{center}
Server
\end{center}

\end{minipage}};
\draw  [fill={rgb, 255:red, 245; green, 166; blue, 35 }  ,fill opacity=1 ][dash pattern={on 4.5pt off 4.5pt}]  (350,290) -- (420,290) -- (420,330) -- (350,330) -- cycle  ;
\draw (385,310) node [anchor=center][inner sep=0.75pt]   [align=left] {\begin{minipage}[lt]{48.08pt}\setlength\topsep{0pt}
\begin{center}
Corrupted\\client
\end{center}

\end{minipage}};
\draw  [fill={rgb, 255:red, 245; green, 166; blue, 35 }  ,fill opacity=1 ][dash pattern={on 4.5pt off 4.5pt}]  (430,290) -- (500,290) -- (500,330) -- (430,330) -- cycle  ;
\draw (465,310) node [anchor=center][inner sep=0.75pt]   [align=left] {\begin{minipage}[lt]{48.08pt}\setlength\topsep{0pt}
\begin{center}
Corrupted\\client
\end{center}
\end{minipage}};

\draw  [dash pattern={on 0.84pt off 2.51pt}]  (395,270) -- (315,290) ;
\draw  [dash pattern={on 0.84pt off 2.51pt}]  (395,270) -- (385,290) ;
\draw  [dash pattern={on 0.84pt off 2.51pt}]  (395,270) -- (465,290) ;

\draw  [fill={rgb, 255:red, 208; green, 2; blue, 27 }  ,fill opacity=1 ][dash pattern={on 0.84pt off 2.51pt}] (360,352) -- (430,352) -- (430,376) -- (360,376) -- cycle ;
\draw (395,364) node [anchor=center][inner sep=0.75pt]   [align=left] {Adversary};

\draw    (395,352) .. controls (371.27,320.48) and (335.11,347.17) .. (315,316) ;
\draw [shift={(315,316)}, rotate = 50.26] [color={rgb, 255:red, 0; green, 0; blue, 0 }  ][line width=0.75]    (10.93,-3.29) .. controls (6.95,-1.4) and (3.31,-0.3) .. (0,0) .. controls (3.31,0.3) and (6.95,1.4) .. (10.93,3.29)   ;
\draw    (395,352) -- (385,330) ;
\draw [shift={(385,330)}, rotate = 62.49] [color={rgb, 255:red, 0; green, 0; blue, 0 }  ][line width=0.75]    (10.93,-3.29) .. controls (6.95,-1.4) and (3.31,-0.3) .. (0,0) .. controls (3.31,0.3) and (6.95,1.4) .. (10.93,3.29)   ;

\draw    (395,352) .. controls (422.49,312.6) and (449.19,360.52) .. (465,330) ;
\draw [shift={(465,330)}, rotate = 126.5] [color={rgb, 255:red, 0; green, 0; blue, 0 }  ][line width=0.75]    (10.93,-3.29) .. controls (6.95,-1.4) and (3.31,-0.3) .. (0,0) .. controls (3.31,0.3) and (6.95,1.4) .. (10.93,3.29);

\draw  [fill={rgb, 255:red, 245; green, 166; blue, 35 }  ,fill opacity=1 ]  (290,58) -- (340,58) -- (340,84) -- (290,84) -- cycle  ;
\draw (315,71) node [anchor=center][inner sep=0.75pt]   [align=left] {\begin{minipage}[lt]{32.77pt}\setlength\topsep{0pt}
\begin{center}
Server
\end{center}
\end{minipage}};

\draw  [fill={rgb, 255:red, 245; green, 166; blue, 35 }  ,fill opacity=1 ]  (350,58) -- (420,58) -- (420,90) -- (350,90) -- cycle  ;
\draw (385,74) node [anchor=center][inner sep=0.75pt]   [align=left] {\begin{minipage}[lt]{48.08pt}\setlength\topsep{0pt}
\begin{center}
Corrupted\\client
\end{center}
\end{minipage}};

\draw  [fill={rgb, 255:red, 245; green, 166; blue, 35 }  ,fill opacity=1 ]  (430,58) -- (500,58) -- (500,90) -- (430,90) -- cycle  ;
\draw (465,74) node [anchor=center][inner sep=0.75pt]   [align=left] {\begin{minipage}[lt]{48.08pt}\setlength\topsep{0pt}
\begin{center}
Corrupted\\client
\end{center}
\end{minipage}};

\draw    (395,117) .. controls (373.27,85.48) and (329.34,128.67) .. (315,84) ;
\draw [shift={(315,84)}, rotate = 67] [color={rgb, 255:red, 0; green, 0; blue, 0 }  ][line width=0.75]    (10.93,-3.29) .. controls (6.95,-1.4) and (3.31,-0.3) .. (0,0) .. controls (3.31,0.3) and (6.95,1.4) .. (10.93,3.29)   ;
\draw    (395,117) -- (385,90) ;
\draw [shift={(385,90)}, rotate = 70] [color={rgb, 255:red, 0; green, 0; blue, 0 }  ][line width=0.75]    (10.93,-3.29) .. controls (6.95,-1.4) and (3.31,-0.3) .. (0,0) .. controls (3.31,0.3) and (6.95,1.4) .. (10.93,3.29)   ;
\draw    (395,117) .. controls (424.49,77.6) and (451.19,125.52) .. (465,90) ;
\draw [shift={(465,90)}, rotate = 117] [color={rgb, 255:red, 0; green, 0; blue, 0 }  ][line width=0.75]    (10.93,-3.29) .. controls (6.95,-1.4) and (3.31,-0.3) .. (0,0) .. controls (3.31,0.3) and (6.95,1.4) .. (10.93,3.29)   ;

\draw  [fill={rgb, 255:red, 126; green, 211; blue, 33 }  ,fill opacity=1 ]  (220, 35) circle [x radius= 60, y radius= 30]   ;
\draw (220,35) node [anchor=center][inner sep=0.75pt]   [align=left] {\begin{minipage}[lt]{52.05pt}\setlength\topsep{0pt}
\begin{center}
Ideal world\\protocol
\end{center}
\end{minipage}};

\draw  [dash pattern={on 0.84pt off 2.51pt}]  (106,35) -- (160,35) ;
\draw  [dash pattern={on 0.84pt off 2.51pt}]  (106,95) -- (160,35) ;
\draw  [dash pattern={on 0.84pt off 2.51pt}]  (315,58) -- (280,35) ;
\draw  [dash pattern={on 0.84pt off 2.51pt}]  (385,58) -- (280,35) ;
\draw  [dash pattern={on 0.84pt off 2.51pt}]  (465,58) -- (280,35) ;

\draw  [fill={rgb, 255:red, 184; green, 233; blue, 134 }  ,fill opacity=1 ]  (50,17) -- (106,17) -- (106,53) -- (50,53) -- cycle  ;
\draw (78,35) node [anchor=center][inner sep=0.75pt]   [align=left] {\begin{minipage}[lt]{35.04pt}\setlength\topsep{0pt}
\begin{center}
Honest\\client
\end{center}
\end{minipage}};
\draw (78,55) node [anchor=north][inner sep=0.75pt]    {$v_{i}$};

\draw  [fill={rgb, 255:red, 184; green, 233; blue, 134 }  ,fill opacity=1 ]  (50,77) -- (106,77) -- (106,113) -- (50,113) -- cycle  ;
\draw (78,95) node [anchor=center][inner sep=0.75pt]   [align=left] {\begin{minipage}[lt]{35.04pt}\setlength\topsep{0pt}
\begin{center}
Honest\\client
\end{center}
\end{minipage}};
\draw (78,115) node [anchor=north][inner sep=0.75pt]    {$v_{i}$};


\draw   (10,175) -- (85,175) -- (85,195) -- (10,195) -- cycle ;
\draw (170,185) node [anchor= center][inner sep=0.75pt]   [align=center] {\begin{minipage}[lt]{100pt}\setlength\topsep{0pt}
\begin{center}
Real parties in the\\ideal world protocol
\end{center}
\end{minipage}};

\draw  [dash pattern={on 0.84pt off 2.51pt}]  (10,235) -- (85,235) ;

\draw (170,235) node [anchor=center][inner sep=0.75pt]   [align=center] {\begin{minipage}[lt]{100pt}\setlength\topsep{0pt}
\begin{center}
A party participates\\ in a protocol
\end{center}
\end{minipage}};

\draw    (10,275) -- (85,275) ;
\draw [shift={(85,275)}, rotate = 180] [color={rgb, 255:red, 0; green, 0; blue, 0 }  ][line width=0.75]    (10.93,-3.29) .. controls (6.95,-1.4) and (3.31,-0.3) .. (0,0) .. controls (3.31,0.3) and (6.95,1.4) .. (10.93,3.29);
\draw (170,275) node [anchor= center][inner sep=0.75pt]   [align=center] {Controls a party};

\draw  [dash pattern={on 4.5pt off 4.5pt}] (10,305) -- (85,305) -- (85,325) -- (10,325) -- cycle ;

\draw (170,315) node [anchor=center][inner sep=0.75pt]   [align=center] {\begin{minipage}[lt]{100pt}\setlength\topsep{0pt}
\begin{center}
Simulated parties in\\ the real world protocol
\end{center}
\end{minipage}};

\draw (170,355) node [anchor=center][inner sep=0.75pt]   [align=center] {Scope of the simulator};

\draw  [fill={rgb, 255:red, 155; green, 155; blue, 155 }  ,fill opacity=1 ][dash pattern={on 0.84pt off 2.51pt}] (10,345) -- (85,345) -- (85,365) -- (10,365) -- cycle ;

\end{tikzpicture}
\caption{Illustration of the simulator}
\end{figure*}
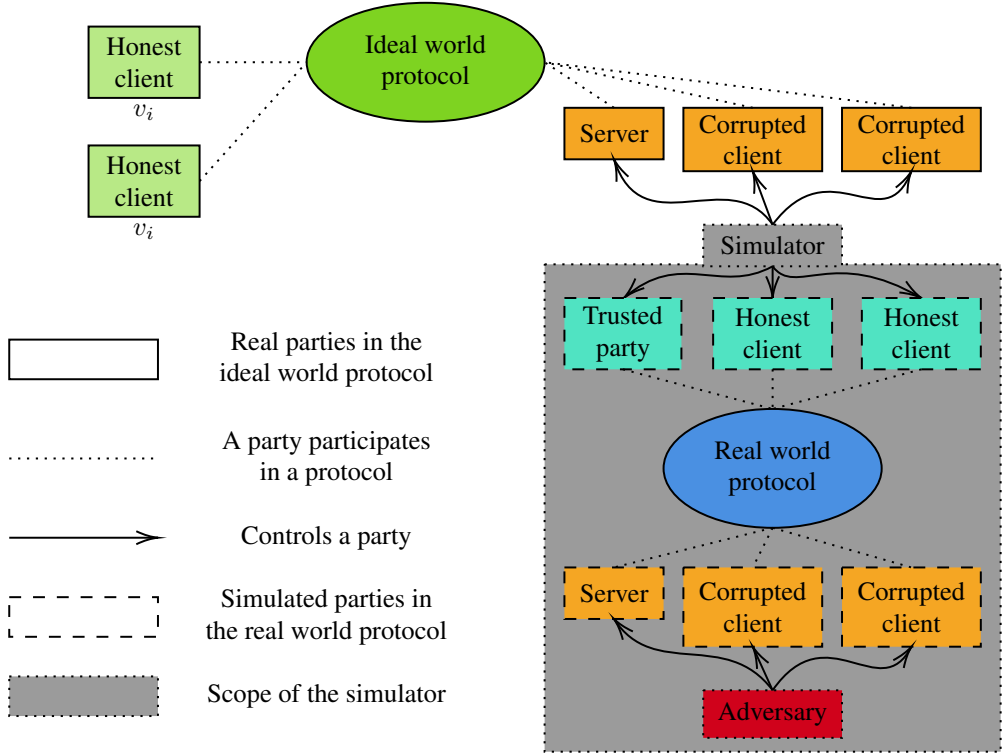

We adopt the simulation-based proof technique to show that the ideal functionality and the real world protocol are computationally indistinguishable. Let $\mathcal{C}_c \subset \mathcal{C}$ denotes the indices of all the clients corrupted by the adversary and $\bar{\mathcal{C}_c} \subset \mathcal{C}$ denotes the indices of the honest clients. Our strategy is to construct a simulator with the following properties
(\cref{fig:simulator_illumination}):

\begin{enumerate}
\item The simulator runs the adversary as a subroutine. 
 \item The simulator executes the real world protocol with the adversary. The simulator plays the role of the trusted party and client $i$ for all $i \in \bar{\mathcal{C}_c}$. The adversary plays the role of the server and corrupted clients $i$ for all $i \in \mathcal{C}_c$.
\item The simulator executes the ideal functionality with the ideal functionality and honest clients. The simulator will play the role of the server and all the clients in  $\mathcal{C}_c$.
\end{enumerate}

We prove that the joint view of the adversary as the simulator's subroutine is computationally indistinguishable from that of a real world execution. The detailed description of the simulator is in \cref{fig:simulator}.

\begin{figure}
    \centering
\fbox{     \begin{minipage}{0.95\linewidth}
    When interacting with the adversary, the simulator follows the real world protocol as the trusted party and client $i$ for all $i \in \bar{\mathcal{C}_c}$, excepts:
    \begin{enumerate}
    \item In \cref{step:key_exchange_response}, for $i \in \bar{\mathcal{C}_c}$, the simulator sends a random vector $\tilde{m}_i\in G^{\ell}$ as $v_i+m_i$ and a random string as the DH key exchange response on behalf of the honest client $i$.
    \item In \cref{step:update_seed}, upon receiving a DH key exchange response with successfully decrypting the encrypted seed from client $i$:
    \begin{itemize}
        \item if $i \in \mathcal{C}_c$, the simulator follows the protocol;
        \item if $i \in \bar{\mathcal{C}_c}$, the simulator adds $i$ to $\mathcal{C}_a$;
    \end{itemize}
    If decrypting the encrypted seed fails, ignore the update.
    
    {\bf Note:} The adversary chooses to aggregate honest clients' inputs whose index is in $\mathcal{C}_a$ and discards rest of the honest clients' inputs.
    
    \item In \cref{step:generating_demasking_vector}, if the trusted party is not expected to generate a unmasking vector, the simulator follows the protocol. Otherwise, the simulator interacts with the real honest clients via the ideal functionality:
        \begin{enumerate}
            \item For each $i \in \mathcal{C}_c$, the simulator sends out a $0$-vector to the ideal functionality on behave of the corrupted client $i$.
            \item The simulator sends $\mathcal{C}_a$ to the ideal functionality.
            \item The simulator receives $V$, which is the sum of all honest clients' inputs, as the server from the ideal functionality.
            \item The simulator sends $(\sum m) + (\sum_{i \in \mathcal{C}_a} \tilde{m}_i)- V$ to the server in the real world protocol on behalf of the trusted party, where $\sum m$ is the running sum maintained by the trusted party.
        \end{enumerate}
        \item In \cref{step:output}, no matter what the adversary outputs in the real world protocol in each role, the simulator outputs the same content as the same role in the ideal functionality.
    \end{enumerate}
    \end{minipage}
    }
            \caption{Simulator for  secure aggregation}
    \label{fig:simulator}
\end{figure}

We now argue that the adversary's views are the same in either the simulation or a real world execution with real honest clients. We will show that by a series of computationally indistinguishable hybrid experiments.

\begin{enumerate}
    \item $\mathbf{Hybrid}_0$: The simulator executes the real world protocol with the adversary. The simulator plays the role of honest clients with their private inputs $v_i$. The adversary plays the role of the server and corrupted clients. This is exactly the real world protocol execution.
    
    \item  $\mathbf{Hybrid}_1$: The same as $\mathbf{Hybrid}_0$, except: 
        \begin{enumerate}
            \item In \cref{step:key_exchange_response}, for $i \in \bar{\mathcal{C}_c}$, the simulator sends a random vector $\tilde{m}_i\in G^{\ell}$ as $v_i+m_i$ and a random string as the DH key exchange response on behalf of the honest client $i$.
                \item In \cref{step:update_seed}, upon receiving a DH key exchange response with successfully decrypting the encrypted seed from client $i$:
                 \begin{itemize}
                      \item if $i \in \mathcal{C}_c$, the simulator follows the protocol;
                      \item if $i \in \bar{\mathcal{C}_c}$, the simulator adds $i$ to $\mathcal{C}_a$;
                \end{itemize}
                If decrypting the encrypted seed fails, ignore the update.
    
                \item In \cref{step:generating_demasking_vector}, if the trusted party is expected to generate an unmasking vector, the simulator will send $(\sum m) + (\sum_{i \in \mathcal{C}_a} \tilde{m}_i) - \sum_{i \in \mathcal{C}_a} v_i$ to the server if no honest clients' response is detected to be tampered with in \cref{step:update_seed}; otherwise sends a uniform random unmasking vector to the server.
        \end{enumerate}
        
        $\mathbf{Hybrid}_0 \approx \mathbf{Hybrid}_1$: \begin{itemize}
                \item The correctness of $\mathbf{Hybrid}_1$ is obvious since the server will get $\sum_{i \in \mathcal{C}_c} (v+m) + (\sum_{i \in \mathcal{C}_a} \tilde{m}_i) -  \left(\sum_{i \in \mathcal{C}_c} m + \sum_{i \in \mathcal{C}_a} \tilde{m}_i - \sum_{i \in \mathcal{C}_a} v_i\right) = \sum_{i \in \mathcal{C}_c \cup \mathcal{C}_a} v$ at the end of $\mathbf{Hybrid}_1$, which is exactly what the server will learn at the end of $\mathbf{Hybrid}_0$.
            
                \item The indistinguishability between  $\mathbf{Hybrid}_0$ and  $\mathbf{Hybrid}_1$ comes from the fact that both  $v_i+m_i$ and $\tilde{m}_i$ are subject to independent uniform distribution over $G^{\ell}$.
        \end{itemize}
        
        \item $\mathbf{Hybrid}_2$: The same as $\mathbf{Hybrid}_1$, except: 
        \begin{enumerate}
            \item The simulator no longer has the inputs of honest clients, but runs the ideal functionality with real honest clients.
              \item In \cref{step:generating_demasking_vector},  if the trusted party is expected to generate an unmasking vector, the simulator interacts with the real honest clients via the ideal functionality:
            \begin{enumerate}
               \item For each $i \in \mathcal{C}_c$, the simulator sends out a $0$-vector to the ideal functionality on behalf of the corrupted client $i$.
               \item The simulator sends $\mathcal{C}_a$ to the ideal functionality.
               \item The simulator receives $V$, which is the sum of all inputs of honest clients, as the server from the ideal functionality.
               \item The simulator sends $(\sum m) + (\sum_{i \in \mathcal{C}_a} \tilde{m}_i)- V$ to the server in the real world protocol on behalf of the trusted party, where $\sum m$ is the running sum maintained by the trusted party.
            \end{enumerate}
        \item In \cref{step:output}, no matter what the adversary outputs in the real world protocol, the simulator outputs the same content as the same role in the ideal functionality.
    \end{enumerate}
    This is exactly the ideal functionality execution with the simulator.
    
    $\mathbf{Hybrid}_1 \approx \mathbf{Hybrid}_2$: 
    The indistinguishability comes from the fact that 
    \begin{itemize}
        \item The ideal functionality will correctly sum up the honest clients' inputs.
        \item The simulator's output in the ideal functionality for each role it plays is identical to the adversary's output in the real world protocol for the same role.
    \end{itemize}
\end{enumerate}

\section{Deployment with Intel SGX}

The  secure aggregation protocol (\cref{fig:protocol_real_world}) we present in \cref{sec:Protocol} involves a trusted party. When deploying this protocol, we use a Intel SGX enclave to play the role of this trusted party. To enforce the honest behavior of this trusted party, we have to ensure the following two security guarantees.

\begin{enumerate}
    \item Confidentiality: the trusted party realized by the enclave shares no information with any other party except what is specified in the protocol.
    \item Integrity: the trusted party realized by the enclave executes the protocol with the correct public parameters (including the group $G$, the vector length $\ell$ and the threshold $t$) without any deviation from the protocol.
\end{enumerate}

In this section, we see how we employ remote attestations and verifiable logs to ensure these properties.

\subsection{Enforce Security with Remote attestations}

Remote attestation technique was originally designed to allow an enclave owner to verify the identity of a trusted binary executed in the cloud. In our use case, there is nothing secret about the code or the initial parameters inside the enclave. Therefore these data can be provided to the clients to allow them play the role of enclave owner and to verify the identity of the trusted binary that plays the role of the trusted party. To be more specific, the extra steps specified in \cref{fig:deployWithSGX} are taken on top of the  secure aggregation protocol in \cref{fig:protocol_real_world} to ensure the honest behavior of the trusted party.

\begin{figure}[h]
    \centering
\fbox{     \begin{minipage}{0.95\linewidth}
\begin{enumerate}
 \setcounter{enumi}{-1}
    \item {\bf Before executing the protocol}, the code of the trusted party running inside the Intel SGX enclave is open sourced in advance along with the hash of the trusted binary, such that the community can exam the code and rebuild the trusted binary running inside the enclave and verify against the claimed hash.
    \item {\bf In \cref{step:key_exchange_request}}, the Intel SGX enclave, playing the role of the trusted party, generates an attestation quote along with each DH key exchange request and sends it to the server. This attestation quote can be used to verify the initial state of the enclave. It consists of the DH key exchange request, the hash of running trusted binary, and the hash of public parameters for the protocol. 
    \item {\bf In \cref{step:send_key_exchange_request}}, the server sends the public parameters used in the protocol and the corresponding attestation quotes to the clients.
    \item {\bf In \cref{step:key_exchange_response}}, upon receiving an attestation quote along with the key exchange request from the server, the client verifies the quote to ensure:
        \begin{enumerate}
            \item the hash of the running trusted binary is the same as the one published with the open sourced code;
            \item the hash of the public parameters provided by the server matches the hash included in the attestation quote. 
        \end{enumerate}
        The client aborts if any of these conditions cannot be verified.
\end{enumerate}
    \end{minipage}
    }
            \caption{Deploying the protocol with Intel SGX enclaves}
    \label{fig:deployWithSGX}
\end{figure}

We follow the standard assumptions of SGX:
\begin{enumerate}
    \item It is infeasible to forge an attestation quote that does not match the running trusted binary and/or the hash of public parameters as the custom payload, but can be verified against Intel's collateral.
    \item It is infeasible to tamper with the trusted binary executed inside the enclave.
    \item It is infeasible to access the data stored inside the enclave except via the predefined APIs.
\end{enumerate}

Under these assumptions and other standard assumptions\footnote{Including: 1. the hash algorithm we use is collusion resistant; and 2. AES is a secure block cipher.}
, clients accept an attestation quote only if:
\begin{enumerate}
    \item the quote is generated by a legitimate enclave;
    \item the enclave is running the predefined code;
    \item the enclave is running with server-claimed parameters;
\end{enumerate}

These arguments jointly assert the enclave is faithfully playing the role of the trusted party. The clients will proceed in the protocol with their private inputs only if they can validate the faithful trusted party. With that said, the server will not hear back from clients unless attestation quotes from a legitimate enclave with correct trusted binary and parameters are forwarded to the clients. 

In addition, the server cannot successfully tamper with the data that is meant to be sent into the enclave, i.e. the DH key exchange response and the encrypted seed. This is because the decryption fails if any of them is modified by the server. Furthermore, the encrypted seed and the response is not accepted by another enclave instance either since it will not have the necessary private randomness to recover the shared key correctly. In summary, the server must use exactly the same enclave during the whole protocol, otherwise it is effectively dropping clients.


\subsection{Updating the Trusted Binary with Verifiable Logs}

Remote attestations allow clients to validate the trusted binary's identity against a hardcoded hash. Such design makes it impossible to update the trusted binary in the future without updating the clients at the same time. To ease the updating process, verifiable logs\cite{verifiablelogs, trillian} can be used to note down any changes made to the code that will run inside an enclave. 

A verifiable log is implemented by a Merkle tree and append-only. Each new record appended to the end of the log is added as a new leaf in the underlying Merkle tree. The hash of the root of the Merkle tree serves as the snapshot of the whole log. An inclusion proof can be generated to demonstrate a record is indeed included in the log. A consistency check can be performed between two snapshots to decide if the corresponding append-only logs are consistent with each other. 

There are several steps to integrate this technique, as detailed in \cref{fig:deployWithverifiablelog}.

\begin{figure}[ht]
    \centering
\fbox{     \begin{minipage}{0.95\linewidth}
\begin{enumerate}
 \setcounter{enumi}{-1}
    \item  {\bf Before releasing the trusted binary}, append the identity and manifest of the trusted binary to the verifiable log if it is not already there.
    \item  {\bf In \cref{step:send_key_exchange_request}}, the server needs to generate an inclusion proof that the trusted binary used in the protocol is included in the latest snapshot.
    \item  {\bf In \cref{step:key_exchange_response}}, the client requests for the latest log snapshot from the server and validates the inclusion proof. The client aborts if the proof cannot be validated.
\end{enumerate}
{\bf Auditing:} Anyone can audit the code running inside the enclave with the following steps:
\begin{enumerate}
    \item Request for the latest log snapshot via the same API.  
    \item Request for all the records (i.e. the trusted binaries) in the log and any of the corresponding source code used for building the trusted binaries to audit.
    \item Check if the source code can be used to build the expected trust binaries. Verify if there is any diversion from the protocol design in the binary. 
\end{enumerate}
    \end{minipage}
    }
            \caption{Deploying the protocol with verifiable logs}
    \label{fig:deployWithverifiablelog}
\end{figure}

Note that both clients and auditors use the same API to request the log's latest snapshot. Therefore the auditors and clients share the same snapshots. Due to the unforgeability of the underlying secure hashes, any logged trusted binary cannot avoid audition without being noticed. On the other hand, clients will only proceed in the protocol only if the trusted binary is logged. In summary, no trusted binary that interacts with clients can avoid audition without getting caught.

With this auditing mechanism in place and sufficient public auditors watching the latest snapshots, the trusted binary can be updated on a regular basis without updating on the client side.

\section{Fixed Point Conversion}

Our secure aggregation protocol works with a finite group. On the other hand, machine learning algorithms operate on real numbers. Hence we need to convert between fixed point and floating point. We observe that plain integer additions and element additions in an integral finite group (e.g. $\mathbb{Z}_{32}$) share the same behavior if there is no wrap-around/overflow. Therefore we cover the gap between real numbers and group elements by using integers as a bridge. A real number is picked as the scaling factor $c$ in advance. Any real number $a$ waiting for aggregation is multiplied by $c$ and rounded to the nearest integer $[ca]$. For the next step an integral finite group ($\mathbb{Z}_{n}$) is picked to simulate the plain integer addition. We map $[-\lfloor n/2 \rfloor, \lceil n/2 \rceil)$ onto $\mathbb{Z}_{n}$ by mapping integer $0, 1, \dots, \lceil n/2 \rceil - 1$ to group element $0, 1,\dots,\lceil n/2 \rceil -1$ and integer $-\lfloor n/2 \rfloor, -\lfloor n/2 \rfloor+1, \dots, -1$ to group elements $\lceil n/2 \rceil, \lceil n/2 \rceil+1, \dots, n - 1$. This conversion allows to support both positive and negative real numbers between $-\lfloor n/2 \rfloor/c$ to $\lceil n/2 \rceil / c$. In summary, a real number is first mapped to an integer before it is finally mapped to a group element in $\mathbb{Z}_n$. 

Note that our protocol in fact does group-element addition on the private inputs. To properly simulate the behavior of integer addition, wrap-round needs to be avoided. In other words, the parties need to estimate the scale of the model updates to aggregate, the desired accuracy to properly pick the parameters including the scaling factor $c$ and the finite group $\mathbb{Z}_n$.

\section{Additional System Design Details}
\label{appendix:system_design}
\subsection{Enforcing Max Concurrency}
To prevent unbounded client participation, the system enforces an upper bound of concurrently participating clients (\textit{C}) for every task based on task configuration. A client can be selected for a task only if number of \textit{active} clients is below the configured threshold. An active client may become inactive for various reasons. The client may have completed execution, or it may be considered dead due to missed heartbeats or execution error. Finally, clients may also be aborted by the server if \textit{staleness} (measured as the difference between current and initial model versions) is higher than a configurable value. 

\subsection{Handling Staleness}

The cost of asynchronous training is staleness of model updates. In this section, we describe how our \asyncfl system tracks and handles staleness. The server model is identified by a \textit{model version}---a non-negative natural number that is incremented every time a new server model is generated. A new server model is generated when $K$ client updates have been aggregated. In an asynchronous FL systems, clients can download a model with an \textit{initial} version, but upload results when the server model has moved to a different \textit{final} version. Recall that we define staleness as the difference between the model version in which a client uses to start local training, and the server model version at the time instance when a client uploads its update.
For each client, the aggregator records initial model version to track staleness. 
Let $s_i$ be the staleness of client $i$ with initial version $V_\text{initial}$ and final version $V_\text{final}$; thus $s_i = V_\text{final} - V_\text{initial}$. We down-weight client $i$'s update using the same scheme as \citet{fedbuff}. Formally, let $w_i$ be the weight of client~$i$ whose staleness is $s_i$, then $w_i\coloneqq 1 / \sqrt{1 + s_i}$. Finally, to bound staleness, the aggregator abort clients whose staleness is larger than a configurable parameter, \textit{maximum staleness}. 

After every server model update, the aggregator aborts clients whose staleness is larger than a configurable parameter, \textit{maximum staleness}. 


\subsection{Switching between SyncFL and AsyncFL}
\papaya highlights that an FL system can support both synchronous and asynchronous training by using client independence, fast model aggregation, high client utilization and asynchronous secure aggregation. These properties improve the performance of both training regime.   

Switching from \syncfl to \asyncfl in our system requires three small changes in behavior: client demand computation, handling of stale clients, and model aggregation.

\textbf{Client demand computation. } In \asyncfl, client demand is computed as $\mathit{concurrency} - \mathit{active\_clients}$. 
However, in a  typical \syncfl round, client demand is high in the beginning of a round, but decreases as clients report results (see Figure~\ref{fig:sync_async_concurrency}). In \syncfl, client demand is computed as $\mathit{concurrency}\cdot(1 + o) - \mathit{completed\_clients}$, where $o$ is the over-selection factor.  

\textbf{Aborting stale clients. } When a server model update is performed in \syncfl, users that are still training are aborted (users may still be training because of over-selection). In \asyncfl, users that are still training continue normally, unless their staleness would exceed \textit{maximum staleness}. 

\textbf{Model Aggregation. } \asyncfl and \syncfl use different model aggregation algorithms.

These three behavior changes are relatively minor. Thus, switching between \syncfl and \asyncfl can be done via a configuration change.

\subsection{Failure Recovery}

Fast recovery and isolated impact from failures help the system minimize model training progress impact. Below we outline mechanisms employed: 

\textbf{Client Routing}. Client requests are routed by selectors using assignment maps (model training task to corresponding aggregator identity) refreshed from coordinator on every report. Upon selector failure or selector having stale assignment map clients retry through a different selector. Failed or stale selector refreshes assignment map on next report to coordinator. 

\textbf{Client Participation}. Coordinator assigns clients to tasks. Upon coordinator failure participating clients are not affected, only for the duration of the recovery no new clients are assigned. Selectors and aggregators wait until a new leader coordinator is elected meanwhile continuing to operate based on last known assignments. After the leader election coordinator enters the recovery period (typically 30s) to rebuild the current assignment map from aggregator reports and then resumes assignments.

\textbf{Task Execution}. Aggregator executes assigned tasks. Upon aggregator failure or unresponsiveness, coordinator detects failures after several missed heartbeats and reassigns all tasks to other aggregators, updates and distributes the new assignment map to selectors. Coordinator detects stale assignments in aggregator reports via sequence numbers and requests to stop executing stale assignments.

\subsection{Edge Training Engine}
The Papaya client is built to be both a hosting platform and an ML framework. An Example Store collects training data in persistent storage and enforces the data use and retention policy. An Executor abstracts model training logic in a general way that supports easily swapping in different ML tasks (data source, model, loss, etc.). The implementation is based on PyTorch Mobile and relies on two features: selective build and the mobile interpreter. Selective build only compiles in ops used by the application to reduce the binary size. The mobile interpreter facilitates efficient cross-platform execution (Android, iOS, Linux) by providing common functionality to save and load model code and parameters, execute forward and backward passes, and optimizer steps.

\end{document}